\documentclass[times,twocolumn,final]{elsarticle}

\usepackage{medima}
\usepackage{framed,multirow}
\usepackage{subfigure}

\usepackage{amssymb}
\usepackage{amsmath}
\usepackage{latexsym}
\usepackage{pgfplots}

\usepackage{url}

\begin{document}

\verso{Rema Daher \textit{et~al.}}

\begin{frontmatter}

\title{A Temporal Learning Approach to Inpainting Endoscopic Specularities and Its effect on Image Correspondence} 

\author[1]{Rema \snm{Daher}\corref{cor1}}
\ead{rema.daher.20@ucl.ac.uk}
\cortext[cor1]{Corresponding author 
  }
\author[1]{Francisco \snm{Vasconcelos}}
\ead{f.vasconcelos@ucl.ac.uk}
\author[1]{Danail \snm{Stoyanov}}
\ead{danail.stoyanov@ucl.ac.uk}

\address[1]{Surgical Robot Vision Group, University College London, Gower Street, London, WC1E 6BT, UK}


\begin{abstract}
Video streams are utilised to guide minimally-invasive surgery and diagnostic procedures in a wide range of procedures, and many computer assisted techniques have been developed to automatically analyse them. These approaches can provide additional information to the surgeon such as lesion detection, instrument navigation, or anatomy 3D shape modeling. However, the necessary image features to recognise these patterns are not always reliably detected due to the presence of irregular light patterns such as specular highlight reflections. In this paper, we aim at removing specular highlights from endoscopic videos using machine learning. We propose using a temporal generative adversarial network (GAN) to inpaint the hidden anatomy under specularities, inferring its appearance spatially and from neighbouring frames where they are not present in the same location. This is achieved using in-vivo data of gastric endoscopy (Hyper-Kvasir) in a fully unsupervised manner that relies on automatic detection of specular highlights. System evaluations show significant improvements to traditional methods through direct comparison as well as other machine learning techniques through an ablation study that depicts the importance of the network's temporal and transfer learning components. The generalizability of our system to different surgical setups and procedures was also evaluated qualitatively on in-vivo data of gastric endoscopy and ex-vivo porcine data (SERV-CT, SCARED). We also assess the effect of our method in computer vision tasks that underpin 3D reconstruction and camera motion estimation, namely stereo disparity, optical flow, and sparse point feature matching. These are evaluated quantitatively and qualitatively and results show a positive effect of specular highlight inpainting on these tasks in a novel comprehensive analysis. 

\end{abstract}

\begin{keyword}
\KWD Deep Learning\sep Endoscopy\sep Specularity Inpainting \sep Computer Vision \sep Temporal GANs
\end{keyword}

\end{frontmatter}


\section{Introduction}
\begin{figure}
    \centering
    \includegraphics[width=\columnwidth]{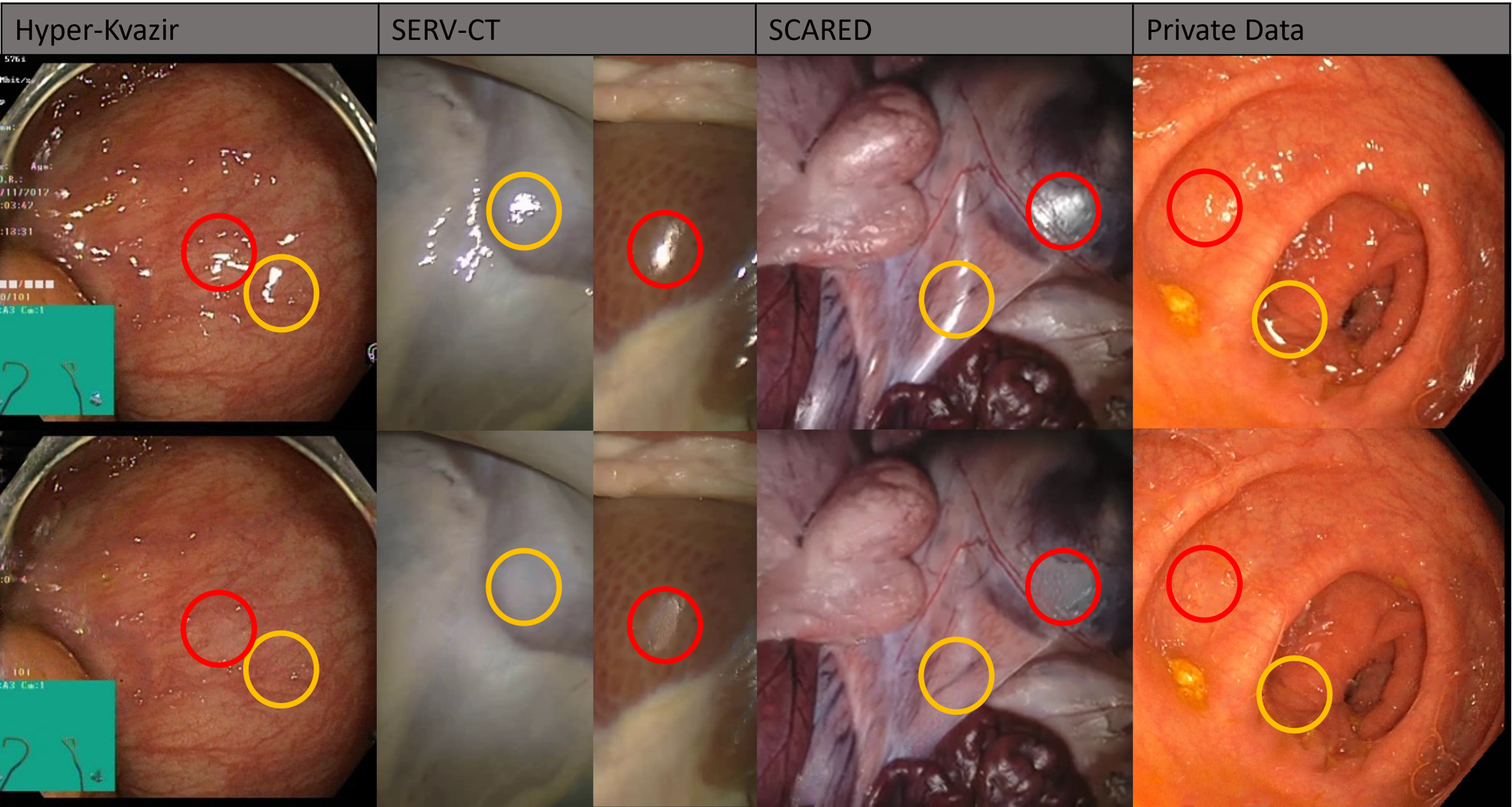}
    \caption{Specular highlight inpainting results with our proposed method. Our model is trained on a portion of the Hyper-Kvasir dataset (upper and lower gastric endoscopy). Provided results are for unseen Hyper-Kvasir images, as well as other datasets without any additional fine-tuning, including in-vivo colonoscopy (private data) and ex-vivo porcine laparoscopy (SERV-CT, SCARED). The yellow circles highlight important inpainted areas where the spatial and temporal components were utilized to recover occluded textures. Red circles show areas where some details were lost with inpainting.}
    \label{fig:motivation}
\end{figure}
\label{sec:Introduction}
Specular highlights in digital images commonly occur with discrete light sources. They present a serious problem in applications that rely on image processing and analysis, such as depth perception, localization, and 3D reconstruction \citep{tao2015depth, ozyoruk2021endoslam}. These highlights not only occlude important colors, textures, and features, but also act as additional features that may be falsely interpreted as characteristic of the scene. 

The effect of specular highlights can be especially harmful when dealing with surgical imaging. Some procedures that are affected by specularities include ossiculoplasty surgery \citep{kalicki2007simple}, dermatological imaging \citep{madooei2015detecting}, cervical cancer screening \citep{kudva2017detection}, laparoscopy \citep{stoyanov2005removing}, endoscopy \citep{ali2021deep}, and cardiac imaging \citep{alsaleh2015automatic}.

This paper focuses on the removal of specular highlights and its effect on other image processing algorithms in the context of minimally invasive surgery (MIS) and diagnostic procedures (MISD).

To be able to diagnose, treat, or take biopsies during MISD, a clear understanding of the surgical scene and the ability to analyze it is of great importance, whether the analysis is done solely visually by the surgeon or with the assistance of computer vision. However, specular highlights highly obscure the visibility of the scene. Specifically, the occlusions caused by specular highlights have a detrimental effect on the surgeons ability to detect anomalies and treat them. According to the study done by \citet{vogt2002making}, surgeons prefer videos with inpainted specular highlights. These highlights also negatively effect the success of numerous MISD computer vision tasks. These tasks include providing a better depth perception, object recognition, motion tracking, 3D reconstruction, localisation, etc \citep{ozyoruk2021endoslam, kaccmaz2020effect}. 

Unlike physical object occlusions (e. g. caused by surgical instruments), specular highlights follow irregular motion and shape patterns, since they change position, disappear, and appear according to tissue properties, light incidence, and camera position. Therefore, several approaches have been developed to specifically target the detection, modelling, and removal of specular highlights. One approach is to segment image regions where specularities are present and remove them from further analysis \citep{menor2016objective, chikkerur2011objective}, however, this may discard useful information from the images. Another approach is to mathematically model the physical properties of these highlights \citep{hao2020photometric}, which can help in estimating 3D shape and deformation of visualised anatomy. This however, is a very complex task that involves accurately modelling light reflection and scattering from the visualised tissue and properties of the camera light source. Finally, the solution that we will focus on is to inpaint these highlights and fill in the missing information. There are two classic approaches to this problem: spatial-temporal patch searching, and diffusion methods. The first one attempts at finding image patches containing the occluded regions in neighbouring video frames where the specularities are not present in the same location \citep{guo2016specular}. Whereas diffusion methods propagate information from the neighbourhood of specular regions within the frame using interpolation \citep{arnold2010automatic}. However, both these classic approaches heavily rely on manually tuned parameters, give low quality results, and fall for local minima. Therefore, learning based approaches have recently been proposed as a more reliable solution \citep{ali2021deep, siavelis2020improved, funke2018generative}.


In this paper, a machine learning approach is used to inpaint specular highlights by training a temporal generative network on gastro-intestinal tract (GI) endoscopy. Endoscopy is a very important procedure under the umbrella of MISD. Endoscopy has been used to diagnose GI tract cancers, perform surgery to treat such problems, or carry out biopsies. The human GI tract cancer has a 63\% mortality rate, which accounts for 2.2 million deaths per year \citep{borgli2020hyperkvasir}
. Being able to diagnose such cancers early on would save many by reducing these numbers significantly. These characteristics make endoscopy an important routine procedure. 

Endoscopic scenes are challenging in computer vision due to their few distinctive texture and geometry features. Another challenge is the uniqueness of endoscopic specular highlights in terms of shape, movement along frames, and conditions for appearance. This makes it hard to learn and model such artefacts, which leads to a lack of ground truth data. Since data is a key component of any learning based approach, synthetic data has been developed, but is still not very realistic \citep{ozyoruk2021endoslam}. In addition to that, with the limited real data in general in the medical field, a clear data shortage arises. All these challenges make this problem different than the usual video inpainting problem.

The available learning based endoscopic specularity inpainting methods have so far only been focused on single frame inpainting. In this paper, a learning based approach with a temporal component will be adapted from general video inpainting and used for the task at hand. This method can take into consideration the relation between consecutive frames as well as neighboring pixel relations. Specifically, an attention based temporal generative adversarial network (GAN) is used to inpaint specular highlights, which in turn enhances endoscopic video streams. 

In addition, to deal with the limited training data and the absence of a ground truth to the occluded textures behind specular highlights, a "pseudo" ground truth was generated. The pseudo ground truth is achieved by translating the specular mask, which decides the regions the model will inpaint, and then removing any overlaps with the original mask. This way, the translated and processed masks now cover regions that have known textures. The paired data that is obtained at this point is called "the pseudo ground truth dataset" since the known textures used are not the original ones that the specular highlights occlude.

Even though our method is trained on GI endoscopy data, experimental evaluation is also extended to laparoscopic ex-vivo porcine data. This is performed to assess the model's effect on datasets with available 3D reconstruction ground truth. The tested datasets include in-vivo GI endoscopic data from the publicly available Hyper-Kvasir dataset \citep{borgli2020hyperkvasir}, colonoscopic data from a private dataset, and laparoscopic ex-vivo porcine data (SERV-CT \citep{edwards2020serv}, SCARED \citep{allan2021stereo}). Some results on the tested datasets are shown in \ref{fig:motivation}. It is worth mentioning that the system was trained only on Hyper-Kvasir data and tested on all 4 datasets. 

To the best of our knowledge, this is the first work in endoscopic highlight removal that employs a deep learning based solution with a temporal component, which effectively exploits occluded texture information from different frames within the video. Another contribution lies in an in-depth, comprehensive analysis of the effect of the same highlight processing pipeline on a diverse range of tasks that are typically used in vision-based localisation, mapping, and reconstruction tasks. We consider in our work comprehensive analysis of the effect of the same highlight processing pipeline on the diverse range of tasks that are typically used in vision-based localisation, mapping reconstruction. The considered tasks in this paper include sparse feature matching, optical flow, camera motion estimation, and stereo disparity calculation. While previous works have studied some of these tasks in isolation \citep{ali2021deep,kaccmaz2020effect,stoyanov2005practical}, we assess the same highlight processing pipeline for all tasks. Furthermore, existing evaluation of feature matching and optical flow relies on synthetic image warping that ignores the characteristic effects of specular highlights in continuous videos such as shape change and disappearance/appearance. To overcome this we use datasets with available geometry ground truth (SCARED, SERV-CT). 

The contributions of the proposed system can be summarised as follows:
\begin{itemize}
    \item A novel solution to specular highlight removal in endoscopic videos. This is achieved with a temporal learning based method as opposed to previous approaches that either ignore the temporal component or rely on parameter heavy patch-based approaches. Our model was adapted from the general-purpose object inpainting approach of \citet{zeng2020learning}.
    \item The generation of pseudo ground truth data for the Hyper-Kvasir dataset that enables effective unsupervised training of our model, as well as quantitative evaluation in unseen images without relying on synthetic image warping.
    \item A quantitative and qualitative comparison of our approach against both learning based and classic non-temporal inpainting approaches, demonstrating the positive effect of temporal modelling.
    \item Analysing qualitatively and quantitatively the effect of inpainting specular highlights on the estimation of stereo disparity, optical flow, feature matching, and camera motion. 
\end{itemize}

The rest of the paper is divided into Section \ref{sec:Related Work}, which includes a discussion of the related work, Section \ref{sec:Proposed System}, which details the proposed system, followed by Section \ref{sec:Experiments}, which describes the experiments performed to achieve the results, that are presented and analyzed in Section \ref{sec:Results}.

\section{Related Work}
\label{sec:Related Work}
This section will depict the work related to handling specular highlights in endoscopy, general-purpose video inpainting methods, and also the effect of specular highlight removal on computer vision tasks in endoscopy.
 
\subsection{Endoscopic Specular Highlight Processing}
Three different approaches have been used to handle specular highlights in endoscopy. The first one and the most simple is to perform highlight detection or segmentation and exclude either image regions or entire frames from further processing \citep{menor2016objective, chikkerur2011objective}. However, such a method results in data loss where occluded texture information is discarded. 

Another approach is to represent specular highlights using a mathematical parametric model. After modelling these highlights, depth as well as motion estimation can be performed from the specular highlight characteristics \citep{hao2020photometric}. However, this does not eliminate the problem of data loss and occluded information that is very important in applications such as polyp detection and feature matching. Additionally, these can be complex models that require priors on the properties of the camera, light source, and the visualised tissue, which can be difficult to obtain in a dynamic environment. 

The third way of handling specular highlights is to inpaint the affected regions with the underlying texture. Some methods rely on an optimization formulation that searches for well matching patches to the occluded region in the rest of the image or in previous frames and uses it as a replacement \citep{guo2016specular}. In this method, motion fields are assumed to be homogeneous and thus do not perform well with complex motions. This method is also considered computationally expensive. An alternative technique is based on diffusion, which smoothly propagates the information in the neighbouring image regions to highlights through interpolation \citep{arnold2010automatic}. 
Both these approaches are limited to local structures and fail to capture the global environment. They also produce low quality results as the missing regions increase in size and have manually tuned parameters. Automated tuning of these parameters has been proposed in \citep{yousaf2014specularity} using a two-layer feed forward neural network. However, the ground truth data is manually chosen based on visual analysis, which adds some inaccuracies. A third inpainting technique relies on data-driven machine learning to automatically generate inpainted specularities such as in  \citep{ali2021deep, siavelis2020improved, funke2018generative}. 

Most learning based approaches to inpainting rely on generative model architectures. Endoscopic videos of the colon, liver, and stomach have been inpainted with an end to end cycle GAN with a self regularization loss \citet{funke2018generative}. The work in \citet{siavelis2020improved} uses GANs to inpaint laparoscopic cholecystectomy images. Cycle GANs have been proposed to restore gastro-esophageal endoscopic specular highlights with an L1 and edge aware loss \citet{ali2021deep}. The model is initialized with the weights of the pretrained model on the Places 2 dataset \citep{zhou2017places}. It is then retrained using a bottleneck approach and tested on endoscopic videos of the oesophagus and pyloric region of the stomach. Finally, \citet{rodriguez2017deep} use a Convolutional Encoder-Decoder network to inpaint GI tract endoscopic specular highlights. However, all these approaches rely on spacial information and do not take advantage of the temporal relation between the video frames.

Reliable ground truth for highlight removal in real surgical video does not exist for any of the analysed procedures, and therefore the evaluation of the above methods has some limitations. Some works only provide visual results for qualitative interpretation \citet{siavelis2020improved} and others perform quantitative evaluations on non-surgical data \citet{rodriguez2017deep}. To perform quantitative evaluation on the targeted surgical domain, one could rely on self-training approaches \citet{funke2018generative} to generate artificial paired data, but this may produce unrealistic images. Alternatively, in \citet{ali2021deep} the authors add randomly placed masks to non-occluded image regions, so that inpainting can be compared against a known tissue pattern. However, random masks do not capture realistic temporal information in continuous video. In this paper, we generate a non-random pseudo ground truth for endoscopic specular highlight masks that is appropriate for continuous video analysis, both during training and evaluation of our models.

\subsection{Temporal Video Inpainting}
Temporal video inpainting has been applied to applications such as re-targeting videos \citep{kim2019deep}, restoring saturated areas \citep{lee2019copy}, and removing objects from videos \citep{tukra2021see}. In fact, all the other state-of-the-art temporal video inpainting methods that will be discussed in this section focus on object or synthetic random shape removal from diverse videos including people, faces, animals, vehicles, and other common objects. The aim of these methods is to create an inpainting technique that works on any application. However, in our application domain more challenges need to be addressed and taken into consideration. These include the discontinuous movement of specular highlights throughout the videos due to the composition of the surfaces of the GI environment and the angle of the incident light. Another challenge is related to the few distinctive features, colors, and textures, as well as the absence of geometric structures in endoscopic videos. Most importantly, video inpainting methods have benchmark annotated ground truth datasets that can be used for training and testing, whereas, this is far from the case in the endoscopic field, where data is limited and the ground truth is hard to obtain.

The proposed architectures for temporal inpainting that mostly focus on object or synthetic random shape removal from diverse videos include 3D and 2D CNNs with a temporal component \citet{wang2019video}, which are improved on in \citet{xu2019deep, zhang2019internal} by estimating appearance simultaneously with optical flow. For coherency in the temporal domain, \citet{kim2019deep} use Recurrent Neural Networks and \citet{chang2019free} inpaint free-form videos using temporal shift modules \citep{lin2019tsm} and temporal SN-PatchGAN \citep{yu2019free}. However, all these methods fail to use information from distant frames. And even with the \citet{gao2020flow} extension on the \citet{xu2019deep} method, where they use edge information for flow estimation and add three distant frames, there is still very limited distant frames taken into consideration. 

To address this issue, attention models have been used. \citet{lee2019copy, oh2019onion} assume homogeneous motions and perform a frame-wise processing of videos without any temporal coherency optimization; only post processing is used, which is time consuming and unreliable with high number of artefacts. \citet{zeng2020learning} address these problems by learning a joint Spatial-Temporal Transformer Network (STTN) using multiscale attention module with similar logic as the traditional method that searches for patches spatially and temporally to fill in the specular occlusions. 

There are a few recent methods than build on STTN to generate higher resolution inpainted textures. These include introducing an Aggregated Contextual-Transformation GAN (AOT-GAN) \citet{zeng2021aggregated}, combining 3D CNNs with a temporal shift and align module \citet{zou2021progressive}, and introducing a Deformable Alignment and Pyramid Context Completion Network with temporal attention \citet{wu2021dapc}. Additionally, more complex occlusions can be handled with a Decoupled Spatial-Temporal Transformer with a hierarchical encoder \citet{liu2021decoupled}.


\subsection{Specular Highlight Removal Effect on Computer Vision Tasks in Endoscopy}

\citet{kaccmaz2020effect} analyse the effect of specular highlight inpainting on classification and detection of polyps in colonoscopy. Other works evaluate tasks related to localisation and reconstruction and are more closely aligned with the work in this paper. For example, \citet{stoyanov2005practical} performs a qualitative analysis of the effect of specular highlights on depth estimation and 3D reconstruction. In addition, \citet{ali2021deep} evaluate the effect of inpainting on optical flow and feature matching, providing a quantitative analysis of the results on synthetically generated data. Known geometric and photometric transformations as well as scaling and Gaussian blur were applied to frames to generate a pseudo ground truth. However this ignores the innate characteristics of specular highlight patterns, such as discontinuity, sudden appearance and disappearance, and change in size and shape. Finally, the above proposed methods do not use any temporal component. 

In this paper we analyse the effect of specularity inpainting both qualitatively and quantitatively on stereo disparity, optical flow, feature matching, and camera motion estimation. The disparity evaluation uses the ex-vivo porcine data with depth ground truth (SERV-CT dataset \citep{edwards2020serv}). Effects on optical flow and feature matching are evaluated by using them for relative camera motion estimation, which can be compared against robot kinematics ground truth on the SCARED dataset \citep{allan2021stereo}. 

\section{Proposed Method}
\label{sec:Proposed System}

At the time of creating the proposed system, the system presented by \citet{zeng2020learning} was the state of the art method in temporal video inpainting. In parallel to STTN, \citet{tukra2021see} developed another generative network to fuse spatial and temporal information.

In this paper, the openly available system presented by \citet{zeng2020learning} is adapted to fit the endoscopic specularity application. The adaptation was performed by training STTN on specularity masks with pseudo ground truth instead of object occlusions with continuous but random masks that can be very different in terms of temporal coherency and continuity. The pseudo ground truth dataset had to be generated and processed to create an end-to-end solution. Finally, transfer learning was utilized to generate better results.

Since the proposed model relies heavily on STTN \citep{zeng2020learning}, a summary of its architecture is provided, followed by an explanation and description of the modifications made.

\subsection{STTN Summary}
First video frames $X_1^T := \{X_1,X_2,...,X_T\}$, their corresponding masks $M_1^T := \{M_1,M_2,...,M_T\}$, along with the target frames $Y_1^T := \{Y_1,Y_2,...,Y_T\}$ are used to train the network. This method follows a similar intuition to traditional methods that search for spatial-temporal patches that require inpainting. This model learns to inpaint missing regions by searching spatially and temporally through neighboring $X_{t-n}^{t+n}$ and distant uniformly sampled frames $X_{1,s}^T$ at a rate of $s$. The problem can be formulated as a "Multi-to-Multi" problem, and using the Markov assumption \citep{hausman1999independence} it can be represented as:
\begin{equation}
    p(\hat{Y}_1^T \mid X_1^T) = \prod_{t=1}^T{p(\hat{Y}_{t-n}^{t+n}  \mid X_{t-n}^{t+n}, X_{1,s}^T)}
\end{equation}

A learning based model is needed here to ensure temporal consistency and coherency between all frames. STTN includes spatial-temporal transformer with multiple layers and multiple heads, sandwiched by a frame level encoder from the left side and a decoder from the right. The encoder and decoder are made up of 2D convolution layers and used to encode features from frames and back again.

\begin{figure}[t!]
    \centering
    \includegraphics[width=\columnwidth]{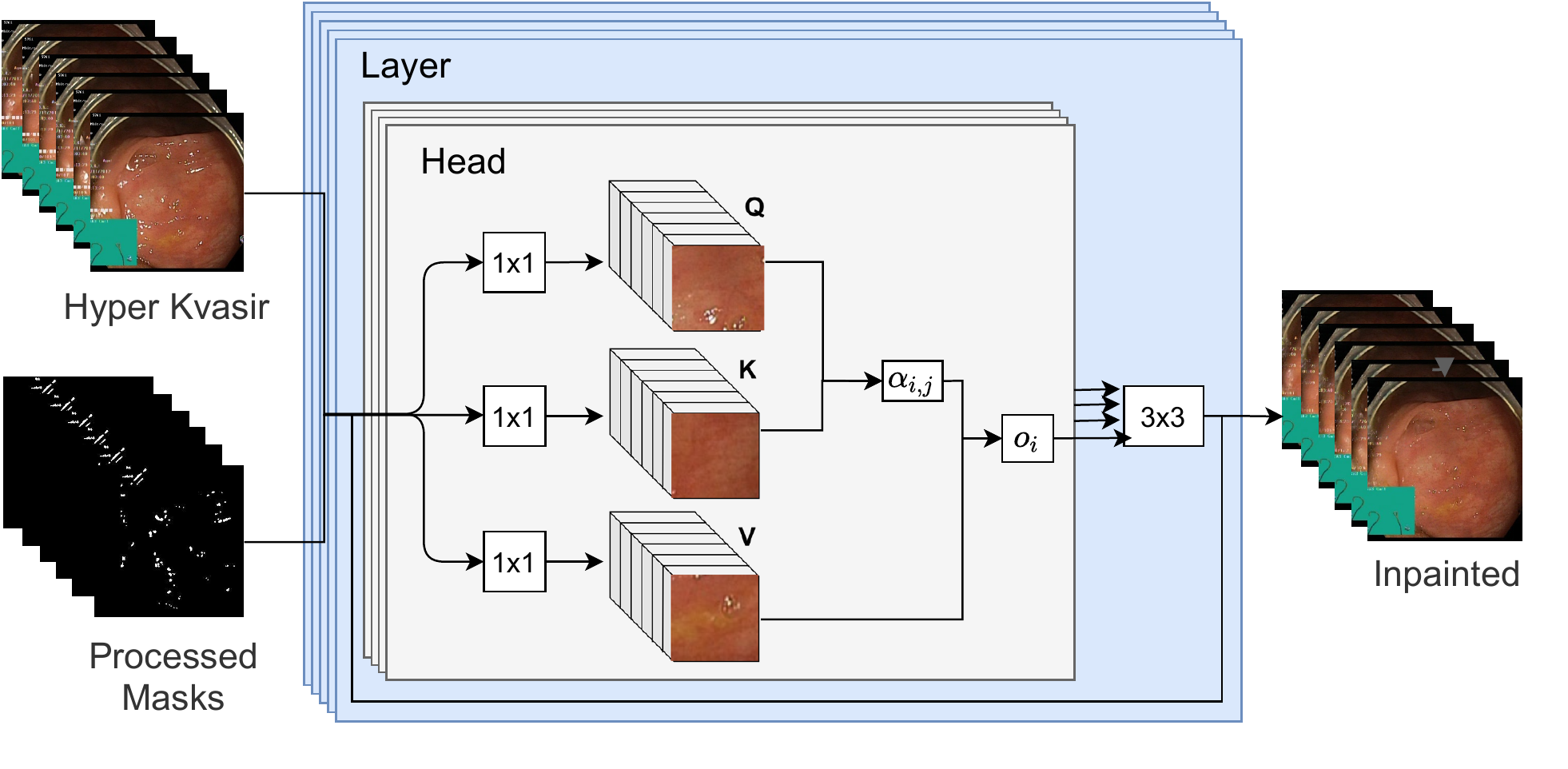}
    \caption{The flowchart of the network architecture of the proposed system with our pseudo ground truth as input. The architecture is a multi-head multi-layer transformer with embedding, matching and attending steps.}
    \label{fig:flowchart2}
\end{figure}

The spatial-temporal transformer as shown in Fig. \ref{fig:flowchart2} has multiple heads responsible for running the transformer across different scales to handle complex motions. For example, large patches that are found in previous frames or in the same frame can be used to inpaint backgrounds and smaller ones can be used to find detailed correspondences in the foregrounds. As for the multiple layers, they help in improving the attention output by taking advantage of updated region features. Empirical studies showed that a number of 8 layers is optimal.

The transformer is made up of three steps, which are embedding, matching and attending. In the embedding stage, the features outputted by the frame level encoder are mapped into query $q_i$ and key-value pair $(k_i, v_i)$ embedding using 1x1 2D convolutions $M_q(f_i), (M_k(f_i), M_v(f_i))$, where $f_i$ are the encoded features.

In the matching stage, which is carried out in every head, spacial patches that will be used to inpaint the specular region are extracted from each frame's $q_i$ and $(k_i,v_i)$, as $p_i^q$ and $(p_i^k, p_i^v)$, respectively. With the patch size being $r1 \times r2 \times c$, the similarity between the $i$-th and $j$-th patches becomes: 
\begin{equation}
    s_{i,j}=\frac{p_i^q \cdot (p_j^k)^T}{\sqrt{r1 \times r2 \times c}}
\end{equation}

The attention weights are then produced for every found patch using the normalized $s_{i,j}$ and a softmax function:

\begin{equation}
    \alpha_{i,j}= 
\begin{cases}
    \exp(s_{i,j})/\sum_{n=1}^N{\exp(s_{i,j})},& p_j \in \Omega\\
    0,              & p_j \in \bar{\Omega}
\end{cases}
\end{equation}
where $\Omega$ and $\bar{\Omega}$ are the unoccluded and missing regions, respectively.

In the attending stage, we can get the output from the query using the attention weights:
\begin{equation}
    o_i=\sum_{j=1}^N{\alpha_{i,j} p_j^v}
\end{equation}

Subsequently, the output from all found patches is pieced together. The results from the various heads get concatenated and inputted into a residual block to maintain frame-wise context.

The optimization objective function used is:
\begin{equation}
    L = \lambda_{hole} \cdot L_{hole} + \lambda_{valid} \cdot L_{valid} + \lambda_{adv} \cdot L_{adv}
\end{equation}
such that L1 losses are used for holes and valid regions with $\odot$ being the element-wise multiplication, which gives the following functions:
\begin{equation}
    L_{hole} = \frac{\| M_1^T \odot (Y_1^T - \hat{Y}_1^T) \|_1}{\| M_1^T \|_1}
\end{equation}
\begin{equation}
    L_{valid}=\frac{\| (1-M_1^T) \odot (Y_1^T - \hat{Y}_1^T) \|_1}{\| (1-M_1^T) \|_1}
\end{equation}
The adversarial loss is also represented as follows:
\begin{equation}
   L_{adv} = -E_{z \sim P_{\hat{Y}_1^T}(z)} [D(z)]
\end{equation}

Empirically, constant values in these functions have been suggested in \citet{zeng2020learning}: $\lambda_{hole} =1, \lambda_{valid} =1$, and
$\lambda_{adv} =0.01$.

\subsection{Modifications}
To adapt STTN to specularity removal in endoscopic videos, some modifications had to be made. First, their model is trained on inpainting random masks in diverse videos, thus the input to the training is the diverse videos and randomly generated masks with frame ensured continuity. However, for specularity removal in endoscopy, endoscopic videos are used along with specularity masks that were preprocessed to create a pseudo ground truth for the occluded regions. A pseudo ground truth generated in an unsupervised manner is needed given that manual ground truth in high volumes is not available and very challenging to obtain.

To generate a pseudo ground truth dataset, the input video streams should be accompanied with specularity masks as well as a pseudo ground truth texture behind these masks. Endoscopic videos are first converted into frames using a 24 frames/second rate. After that, the specularity masks are generated from the frames using the segmentation method proposed by \citet{el2011automatic}. This segmentation technique is based on the Dichromatic Reflection Model (DRM) and makes use of the chromatic characteristics of the specular highlights. Other methods could potentially be applied, but alternatives either have similar performance or are not yet open source such as learning based methods.

\begin{figure}[t!]
    \centering
    \includegraphics[width=0.3\columnwidth]{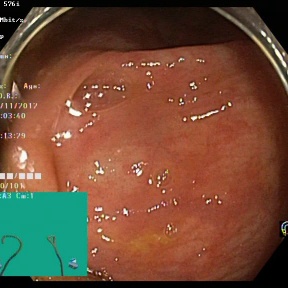}
    \includegraphics[width=0.3\columnwidth]{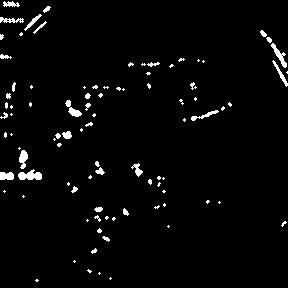}
    \includegraphics[width=0.3\columnwidth]{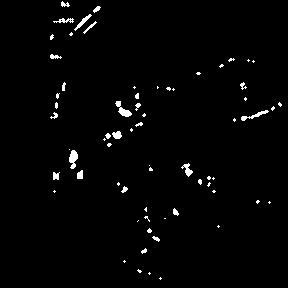}
    \caption{The original video frame (left) is segmented generating a specularity mask (middle), which is then processed (right) for the purpose of creating a pseudo ground truth.}
    \label{fig:masks}
\end{figure}

After segmentation, the masks are processed by translating them in a way that would make them cover up visible/unoccluded texture that acts as a pseudo ground truth. The resulting masks are denoted as $Masks_{Trans}$. To do that, the specularities were first translated to the right and any overlap between their current and old specularity locations was not taken into consideration as shown in Fig. \ref{fig:masks}. After the masks are translated and processed, they move to a position that was originally without specularity and thus the texture behind the generated masked regions would be known. That way paired data is achieved, but since it is not the real occluded textures of the original specularities, we call these textures "pseudo ground truth" making the name of the dataset "the pseudo ground truth dataset". 

\begin{figure}[t!]
    \centering
    \includegraphics[width=\columnwidth]{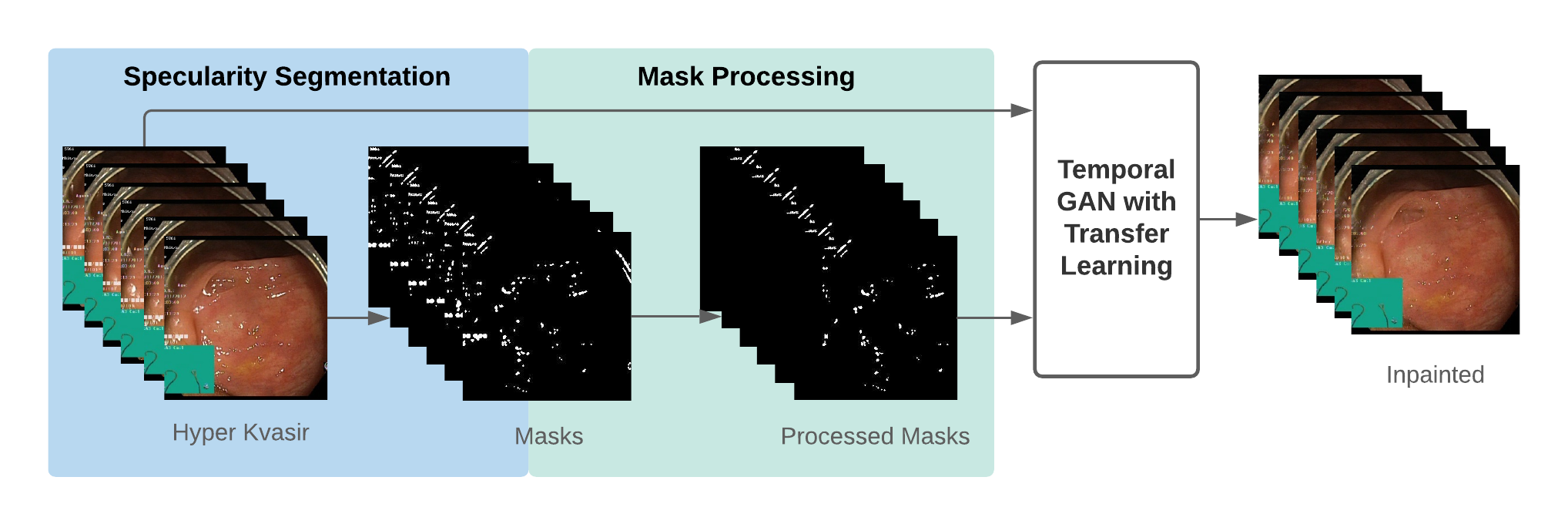}
    \caption{The flowchart of the proposed system is comprised of a specularity segmentation step, where specularity masks are generated from the input video dataset. After that, the masks are processes (moved to new unoccupied locations) to create a pseudo ground truth dataset. The pseudo ground truth along with the original video of the Hyper-Kvasir dataset are used to train a temporal GAN, which is initialized by a model trained on random masks. The output seen on the far right is the inpainted frames when the input mask contains the original specularity locations, which is done in our qualitative analysis.}
    \label{fig:flowchart}
\end{figure}

Thus, the input training masks were modified from randomly generated masks of the STTN baseline method to segmented and processed specularity masks, $Masks_{Trans}$. Moreover, training on temporally continuous random masks with endoscopic videos was used as an initialization phase followed by fine-tuning on $Masks_{Trans}$. This transfer learning technique was performed to help the model obtain an a more accurate initial guess and reach a global minimum. The masks, $Masks_{Trans}$, were also dilated using an ellipse morphological structure and $8 \times 8$ iterations to cover the dark rings that usually occur around specularities. In addition, the input frames were cropped and resized to $288 \times 288$ as opposed to the input of $432 \times 240$ used by STTN. That is because having a uniform square size to the different video frames in our pseudo ground truth dataset made it more efficient to deal with the data. Thus, a change also had to be made to the sizes list of the patches to be searched for, which in turn effected feature sizes. The proposed system then becomes as shown in Fig. \ref{fig:flowchart} and \ref{fig:flowchart2}.

\section{Experiments}
\label{sec:Experiments}
For training and testing, an NVIDIA V100-DGXS was used. The endoscopic videos used for data generation are the videos of the Hyper-Kvasir dataset \citep{borgli2020hyperkvasir}. The Hyper kvasir videos are raw videos of routine clinical examinations of the GI tract using endoscopy. This dataset is made up of 373 videos with 889,372 frames. 343 of those were used for training and 30 (8.7\%) were used for testing, which is close to the split used by STTN \citep{zeng2020learning}. Even though STTN uses 4,603 videos, their frames only reach 681,450 making our pseudo ground truth dataset of a similar and even larger size. In addition, our task of specular highlight inpainting has very narrow video content and similar features as opposed to the various video stream content in the datasets STTN used such as Youtube-VOS \citep{xu2018youtube} and DAVIS \citep{caelles20182018}.

To evaluate results and since paired data is available (data with pseudo ground truth), Peak Signal to Noise Ratio (PSNR) and Mean Square Error (MSE) are generated at cropped specularity regions of $Masks_{Trans}$. Since these metrics compare a pair of frames, the pair used here is the inpainted frame and its pseudo ground truth. Then, the average of the metrics along all the frames in a video is calculated. The Structural Similarity Index Mean (SSIM) could not be used, because it takes into consideration the relations within windows in a frame, whereas, in the case at hand only the cropped parts need to be evaluated. Even if SSIM is computed on the whole image and then only the values for the cropped pixels are taken, the values will be too close to each other and very hard to analyze. That is why, only pixel-based metrics are used.

This evaluation process is applied on all testing videos for every trained model. After that, another averaging step is performed between the values generated from all testing videos in order to get one value for every model, which allows for a quantitative analysis between models.

To assess the generalizability of our system, various datasets where qualitatively evaluated including in-vivo data of gastric endoscopy (Hyper-Kvasir and private data from another hospital) and ex-vivo porcine data (SERV-CT, SCARED). It is is worth mentioning that the model was not fine-tuned on these datasets because they are limited in size.

In addition to that, further analysis is performed on the effect of inpainting endoscopic specular highlights on disparity and optical flow estimation along with feature matching. For the disparity estimation analysis, the ex-vivo porcine SERV-CT dataset is used, which includes accurate ground truth disparities and an evaluation system. The disparity estimation method used is a recent open source learning based method called Stereo Transformer proposed by \citet{li2020revisiting}. 

As for the feature matching analysis, Scale-invariant feature transform (SIFT) detector proposed by \citet{lowe1999object} was used to generate keypoints and descriptors. After that, the Fast Library for Approximate Nearest Neighbours (FLANN) was utilized for matching \citep{muja2009fast}. The ratio test as per \citet{lowe1999object} was then used to remove low quality matches. From this, a visual analysis was performed using our dataset, but with the original specularity masks. It is worth mentioning that these computer vision blocks are used in this procedure to illustrate the importance of inpainting in feature matching; however, they can be replaced by any other detector, matcher, and outlier removal methods.

Moving on to the optical flow analysis, the optical flow estimation network Flownet2.0 was used; it was proposed by \citet{IMKDB17} and implemented by \citet{flownet2-pytorch}. This method can also be replaced by any other optical flow estimation technique to illustrate the importance of inpainting. For the visual assessment our dataset was also used with the original specularity masks as well as the ex-vivo porcine data with 3D scene ground truth (SCARED dataset).

Along with the visual evaluation, a quantitative assessment was performed to analyze the effect of inpainting specular highlights in endoscopy on sparse feature matching and optical flow. A direct evaluation of optical flow and feature matching was not possible due to the lack of ground truth data. Thus, the matching results, from sparse feature matching or optical flow, were used to generate pose data, which was evaluated with the ground truth pose provided by the SCARED dataset. The SCARED dataset consists of ex-vivo porcine data with 3D scene ground truth. Ex-vivo data was used since available endoscopic data does not have any ground truth information for evaluation; that way, we also assess the extension of our model to other types of surgical data. 

To generate and evaluate pose data from the matches, the essential matrix was generated based on the five-point algorithm solver \citep{nister2004efficient}, which relies on the RANSAC algorithm \citep{fischler1981random}. After that, the pose was recovered by decomposing the essential matrix and then verifying possible pose hypotheses using the cheirality check \citep{nister2004efficient}. For this pipeline, an implementation by Gyanesh Malhotra\footnote{\url{https://github.com/gyanesh-m/Monocular-visual-odometry}}. After that, relative poses were evaluated with the ground truths and compared between original and inpainted frames. Due to the noise in the ground truth pose information of the SCARED datasets that can appear more dominant with small movements, enough movement between frame pairs had to be ensured. To do that, a moving 20 frames window was used along the sequence to collect frame pairs and evaluate them. This resulted in 715 frame pairs to test on.

It is worth mentioning that the effect of inpainting on feature matching or optical flow and consequently pose estimation might be more visible with endoscopic datasets since the SCARED dataset has very steady and continuous motion and much less artefacts that might degrade flow estimation or feature matching. In addition, our model was trained on endoscopy and would naturally perform better on similar testing data. It is also important to note that even though pose estimation is used to assess feature matching and optical flow estimation, it will not be as effected by inpainting as feature matching and optical flow. This is because the RANSAC method used to estimate pose removes outliers, which can make it somewhat robust to some of the artefacts in surgical video streams.

\section{Results and Discussions}
\label{sec:Results}
In this section, results for various models trained and tested is first evaluated and discussed, followed by an ablation study for the importance of the temporal component of our system in the application of specular highlight removal in endoscopy.

The applications of our system are numerous and include the improvement of 3D reconstruction, feature matching, and the estimation of disparity and optical flow. To show our systems effect on some of these applications, we first started with an analysis for disparity estimation, followed by feature matching, and ending with optical flow.

\subsection{Model training}
In our experiments, several models were generated and compared to reach the best outcome. The models included $Model_{S,R}$, which is trained from scratch on Hyper-Kvasir with temporally continuous random masks. This training is similar to that of the baseline method, STTN, where diverse videos with random continuous masks are used for training. This model was used to initialize $Model_{T,C}$, which relies on transfer learning and is trained on our pseudo ground truth dataset with $Masks_{Trans}$. To show the importance of transfer learning, another model was generated $Model_{S,C}$  that was trained from scratch on our pseudo ground truth dataset. 

\begin{figure}[t!]
    \centering
\subfigure[]{    
    \includegraphics[trim=2.1cm 0 0 0, clip, width=0.48\columnwidth]{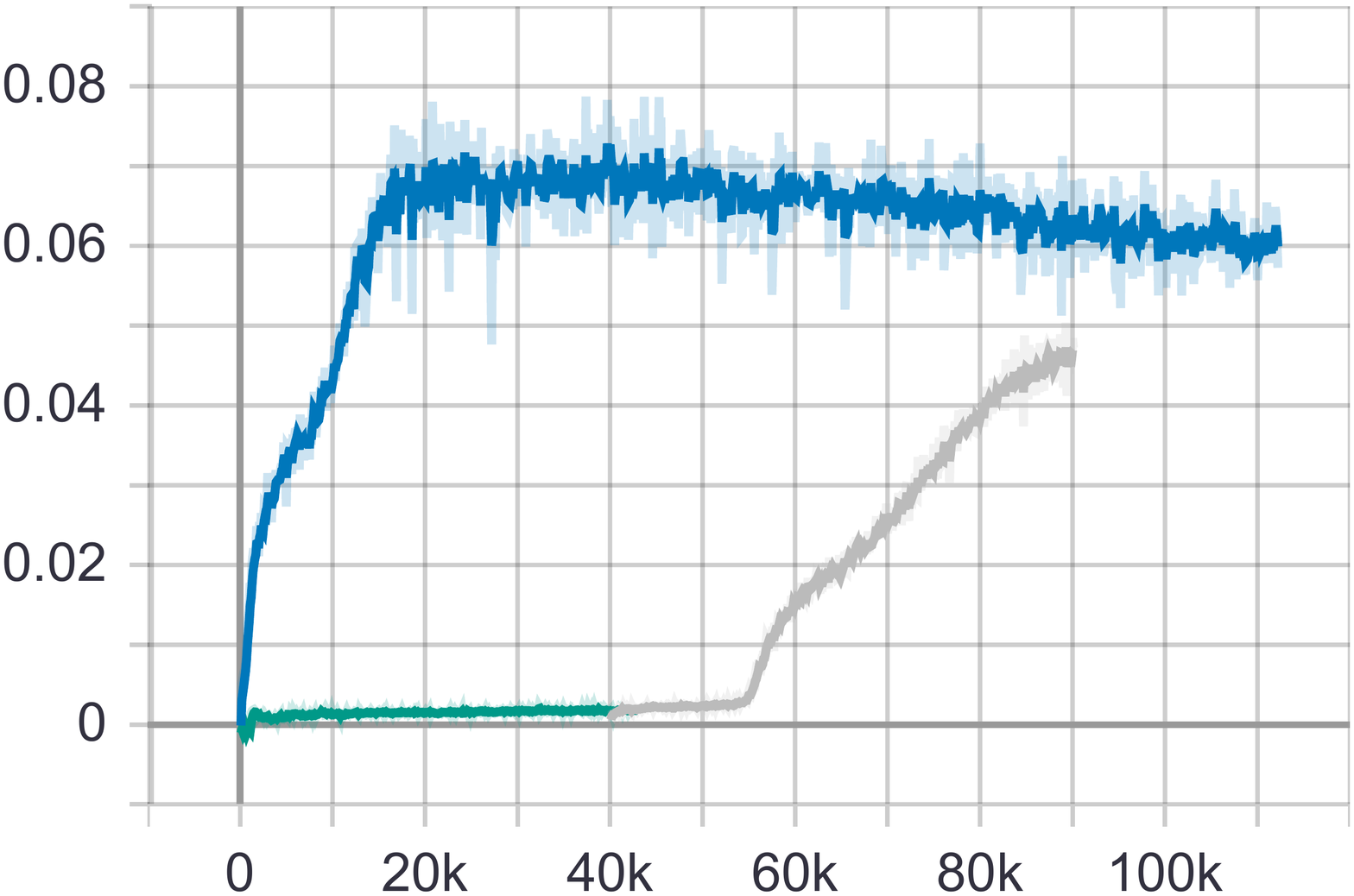}}
\subfigure[]{
    \includegraphics[trim=2.1cm 0 0 0, clip, width=0.48\columnwidth]{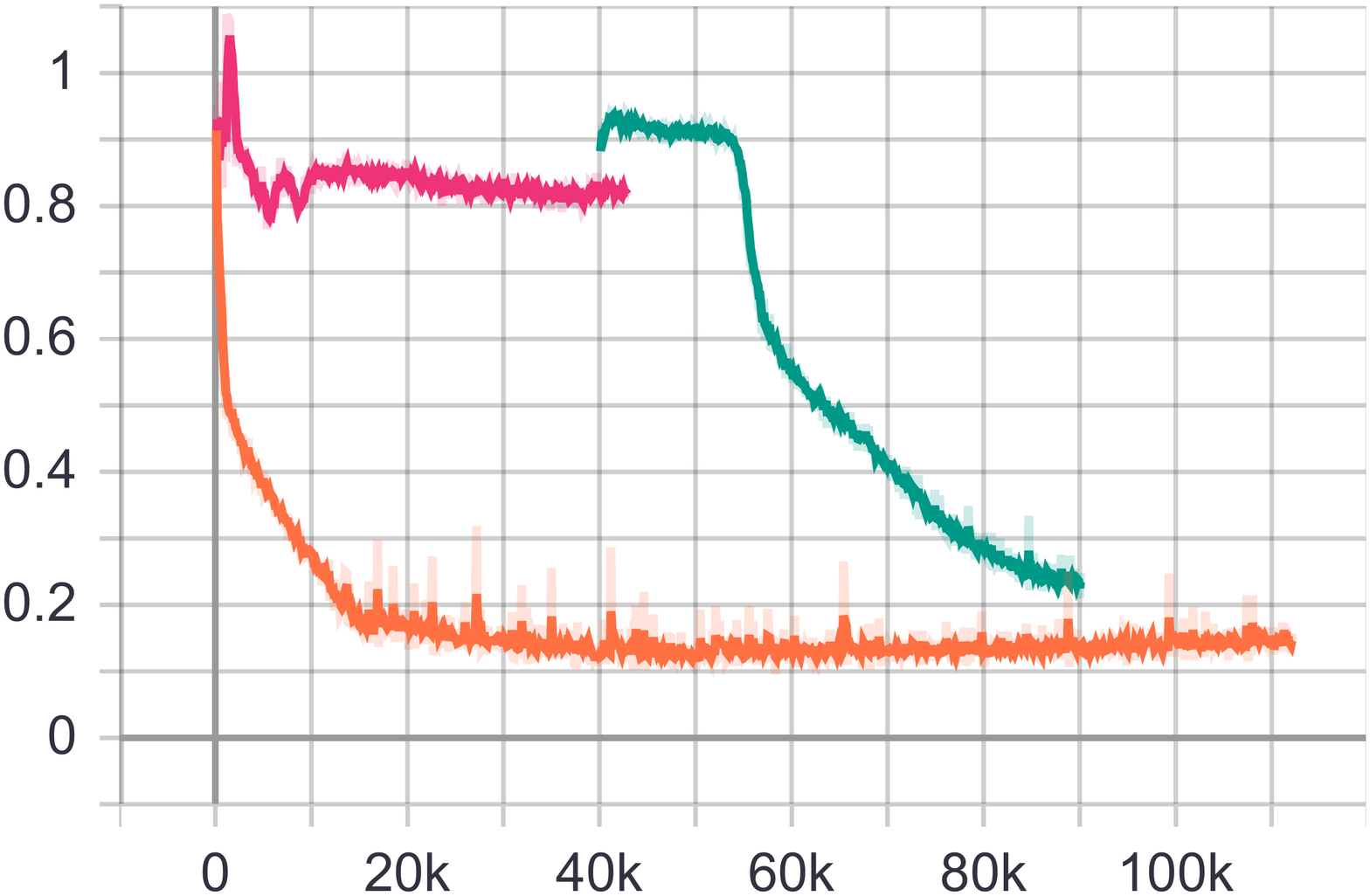}}
    \caption{(a) The plots represent $L_{adv}$ generator losses, where green corresponds to $Model_{S,R}$, grey to $Model_{T,C}$, and blue to $Model_{S,C}$. The plots show that $Model_{T,C}$ and $Model_{S,C}$ outperform $Model_{S,R}$. 
    (b) The plots represent the real discriminator losses, where pink corresponds to $Model_{S,R}$, green to $Model_{T,C}$, and orange to $Model_{S,C}$. The plots show that $Model_{T,C}$ and $Model_{S,C}$ outperform $Model_{S,R}$.}
    \label{fig:discriminator}
\end{figure}

The generator loss $L_{adv}$ for these models can be shown in Fig. \ref{fig:discriminator} (a). The real discriminator losses can also be shown in Fig. \ref{fig:discriminator} (b). 

\begin{figure}[t!]
    \centering
    \includegraphics[width=\columnwidth]{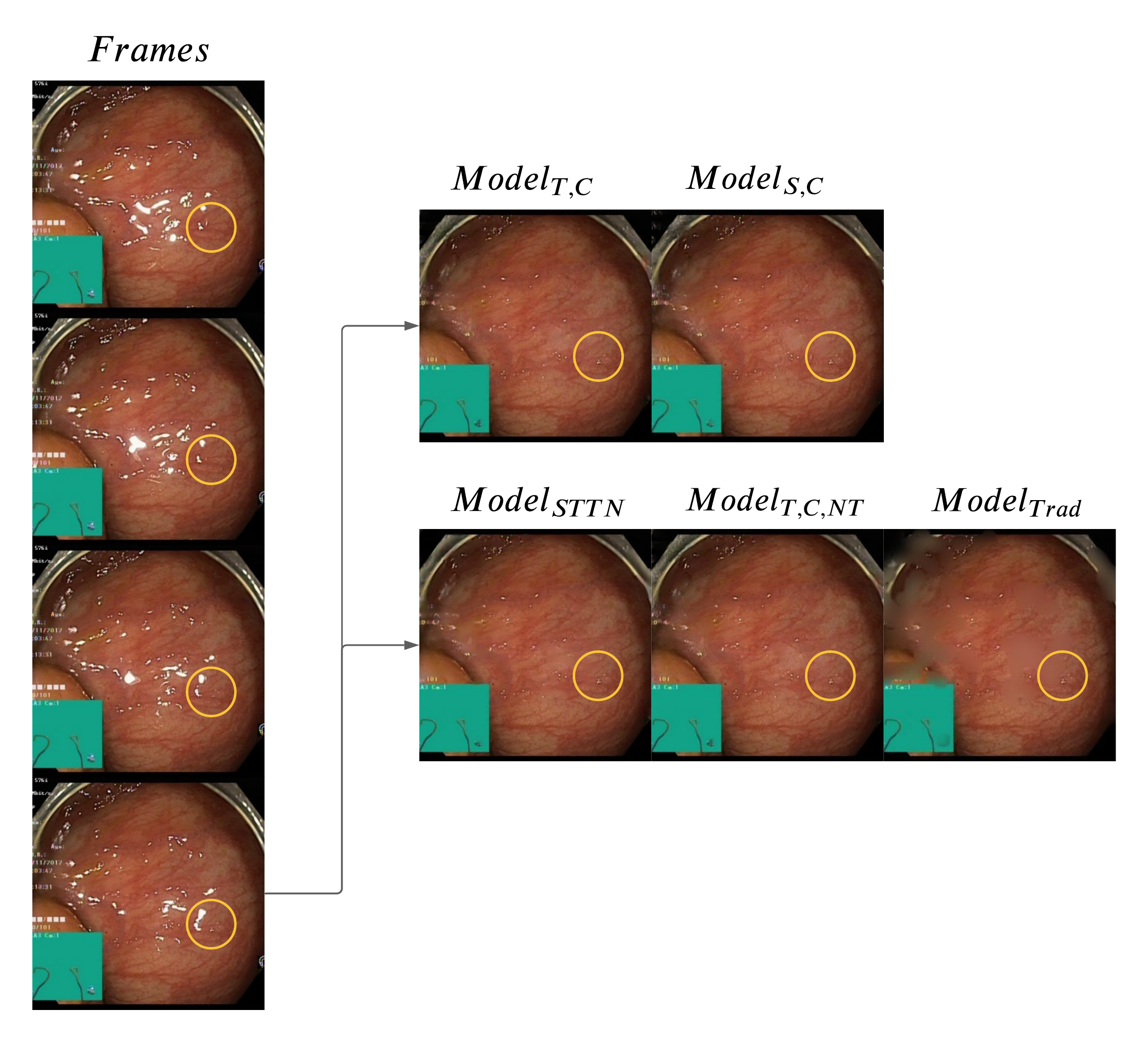}
    \caption{Four consecutive frames are inpainted by $Model_{T,C}$ and $Model_{S,C}$, $Model_{STTN}$, $Model_{T,C,NT}$, and $Model_{Trad}$. The yellow circles highlight a region where specularities increase in size. It is shown that $Model_{T,C}$ outperformed other models and was able to temporally recover occluded regions that were not occluded in earlier frames.}
    \label{fig:scratchVis}
\end{figure}

Given that in GANs the losses are a trade off, the optimal loss is reached at a plateau somewhere around the middle between 0 and 1. From these plots (Fig. \ref{fig:discriminator} (a) and (b)), it can be seen that $Model_{S,R}$ on its own plateaus very close to the edges of this range $(0,1)$, which indicates that it does not perform well. However, with transfer learning, $Model_{T,C}$ losses converge more to the middle indicating better results. To evaluate the importance of transfer learning, a comparison between $Model_{T,C}$ and $Model_{S,C}$ is also carried out. However, it is not very clear from the losses which of the two gives better results, it shows that they both converge between 0 and 1 and that $Model_{T,C}$ plateaus more towards the middle, which is not enough to compare. That is why, quantitative and visual analysis is also carried out in Fig. \ref{fig:scratchVis} and Table \ref{tab:scratchTab}.

\begin{table}[t!]
    \centering
    \begin{tabular}{|c|c|c|c|}
    \hline
    Models        & Iterations  & $\blacktriangle PSNR_{mean}$  & $\blacktriangledown MSE_{mean}$\\
    \hline
    \hline
    $Model_{S,C}$ & 20,000      &	27.138                      &	164.542\\
    \hline
    $Model_{S,C}$ & 110,000      &	28.896                      &	120.613\\
    \hline
    $Model_{T,C}$ & 50,000      &	29.200                      &   115.181\\
    \hline
    $Model_{T,C}$ & 60,000      &	29.435                      &	107.565\\
    \hline
    $Model_{T,C}$ & 80,000      &	29.181                      &	111.201\\
    \hline
    $Model_{T,C}$ & 90,000      &	\textcolor{red}{29.542}     &	\textcolor{red}{104.719}\\
    \hline

    \end{tabular}
    \caption{The $PSNR_{mean}$ and $MSE_{mean}$ values for $Model_{S,C}$ and $Model_{T,C}$ at different iterations. $PSNR_{mean}$ values are better as they increase, which is shown by the $\blacktriangle$ and the opposite can be said about the $MSE_{mean}$ values}
    \label{tab:scratchTab}
\end{table}

$Model_{T,C}$ was saved at 90,000 iterations, which was decided on the basis of the inflection point where the loss plots plateau as well as on visual and quantitative analysis. Similar logic was carried out for $Model_{S,C}$ which was used at iteration 110,000 and even though that was not the inflection point, it did give slightly better results than the results at 20,000 iterations visually and quantitatively. 

The quantitative analysis for various iterations of the models can be seen in Table \ref{tab:scratchTab}. From Table \ref{tab:scratchTab}, it can be seen in red that the best performing model according to $PSNR_{mean}$ and $MSE_{mean}$ is $Model_{T,C}$ at 90,000 iterations, indicating the advantage of transfer learning in this application. In addition, in Fig. \ref{fig:scratchVis}, four consecutive frames are inpainted by both $Model_{T,C}$ and $Model_{S,C}$. From these figures, it can be seen that the $Model_{T,C}$ learns better temporal inpainting; in the yellow circles the red veins are much more prominent and fine-grained for $Model_{T,C}$, which means they have been learnt from previous frames. These images also suggest that even if the quantitative analysis doesn't show a huge increase in $PSNR_{mean}$ and $MSE_{mean}$ values, it can make a difference in terms of accurate vein depiction. This is due to the fact that the textures in endoscopic frames are very close to each other making any small change count in showing the details better.

Having chosen $Model_{T,C}$ for this system, a comparison between this model and other methods is important. However, at the time of creating this system, learning based methods for endoscopic specularities were not yet available as open source. That is why, a comparison between this method and the traditional method proposed by \citet{arnold2010automatic}, which uses diffusion based inpainting, was carried out. This traditional model will be denoted by $Model_{Trad}$. 

The visual results of the traditional method compared to ours can be seen in Fig. \ref{fig:scratchVis}, where the inpainted regions are blurry and have a green shade, which is very far off the real texture. In addition in the yellow circle it can be seen that no veins were generated with mostly one color dominating the region. This can also be seen in the quantitative results in Table \ref{tab:temp}, where our system gives much better results. These results make sense since the diffusion method smoothes pixel values into the empty regions instead of accurately depicting the missing features as can be done using learning based methods and specifically through the temporal component.


\begin{table}[t!]
    \centering
    \begin{tabular}{|c|c|c|}
    \hline
    Models              & $\blacktriangle PSNR_{mean}$  & $\blacktriangledown MSE_{mean}$\\
    \hline
    \hline
    $Model_{T,C}$       &	\textcolor{red}{29.542}     &	\textcolor{red}{104.719}\\
    \hline
    $Model_{STTN}$      &	28.683                      &	119.541\\
    \hline
    $Model_{Trad}$      &	19.909                      &	895.222\\
    \hline
    $Model_{T,C,NT}$    &   29.284                      &	112.717\\
    \hline
    \end{tabular}
    \caption{The $PSNR_{mean}$ and $MSE_{mean}$ values for $Model_{STTN}$, $Model_{T,C}$, $Model_{T,C,NT}$, and $Model_{Trad}$. $PSNR_{mean}$ values are better as they increase, which is shown by the $\blacktriangle$ and the opposite can be said about the $MSE_{mean}$ values}
    \label{tab:temp}
\end{table}

\subsection{Ablation study}

An ablation study is performed to observe the effect of the temporal component on the output. We compare our proposed model in its original form against the same model but only considering a single frame input (i.e. it does not consider temporal component). This experiment is denoted as $Model_{T,C,NT}$. The visual and quantitative results can be seen in Fig. \ref{fig:scratchVis} and Table \ref{tab:temp}, respectively. It is shown that excluding the temporal component not only gives worse quantitative results, but also misses details that can be obtained from previous frames. To elaborate, if one follows the yellow circle throughout the original frames in Fig. \ref{fig:scratchVis}, it is clear that the region gets more occluded with every frame, meaning that a temporal component should be able to take advantage of unoccluded regions from previous frames and exploit it during inpainting. This advantage can be seen when one also compares the output inside the yellow circles for $Model_{T,C,NT}$ and $Model_{T,C}$, where a prominent vein is blurred in $Model_{T,C,NT}$ while its details are visible in $Model_{T,C}$. 

The original STTN model \citep{zeng2020learning}, trained on Youtube-VOS, $Model_{STTN}$, was also tested on our pseudo ground truth dataset to see the effect of our training methodology. The visual and quantitative results of these models can be seen in Fig. \ref{fig:scratchVis} and Table \ref{tab:temp}, respectively. The results show a visual improvement of our method in the yellow circle of Fig. \ref{fig:scratchVis}.

\subsection{Generalizability}

We performed an analysis of the generalizability of our system, without any additional fine-tuning. Our system trained on Hyper-kvasir data was tested on 3 additional datasets with different characteristics. The results of sampled frames can be visualised in Fig. \ref{fig:motivation} (additional results are provided in the supplementary material). Another private dataset was used that, similarly to Hyper-Kvasir, includes in-vivo data of gastric endoscopy, however it was acquired on a different site and with different equipment. This data was collected in Clinico Lozano Blesa hospital using Olympus Europe endoscopic system, whereas Hyper-Kvasir was performed in Bærum Hospital using standard equipment from Olympus Europe and Pentax Medical Europe. In general, the results of this data were satisfactory, where veins were uncovered under specularities as seen in yellow in Fig. \ref{fig:motivation}. However, sometimes smaller specularities are inpainted without much details as seen in red in Fig. \ref{fig:motivation}. This means that the model was able to generalize to other hospitals and equipment, however, the result are not as good as the dataset used in training.

The other two datasets (SERV-CT, SCARED) are of ex-vivo porcine data acquired with a laparoscope. SERV-CT only has keyframes, which makes the temporal component of our system not utilized fully. However, it still gave good results as can be seen in yellow in Fig. \ref{fig:motivation}. The red circles show an error that occurs due to the detection system's inability to detect saturation. As for the SCARED dataset, the frames are closer, but since it is also ex-vivo data, the colors and textures are different than the trained dataset, which makes the results not detailed as seen in red. However, we can still see some successful inpaintings in the yellow circles. To conclude, our model can be generalized to ex-vivo data with lower quality results especially in regions where colors are different than those in in-vivo endoscopy. 

\subsection{Stereo Disparity}
\begin{figure}[t!]
    \centering
    \includegraphics[width=\columnwidth]{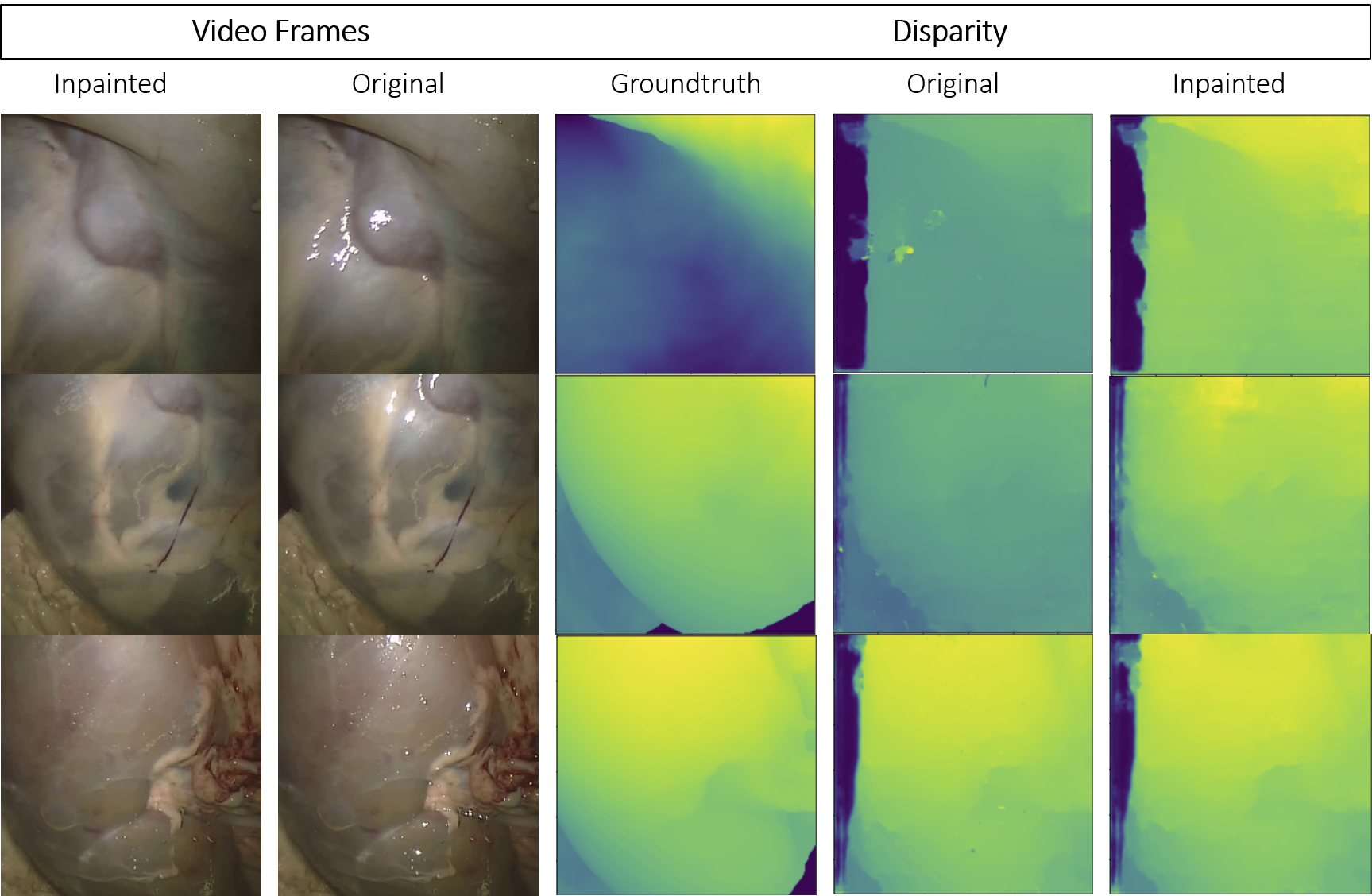}
    \caption{The ground truth and estimated disparity based on original as well as inpainted frames of $Model_{T,C}$. Even though SERV-CT frames are keyframes, making the temporal component of the model not used to its full potential, the disparity from inpainted frames shows improvement over that from original frames.}
    \label{fig:disparity}
\end{figure}

In this section, we evaluate the effect of inpainting endoscopic specular highlights on stereo disparity calculation. The SERV-CT evaluation system and dataset, which includes ground truth disparities, are used. As shown in Fig. \ref{fig:disparity}, the disparity generated is closer to the ground truth after inpainting the SERV-CT data using $Model_{T,C}$. Some spurious artifacts in the dispartity maps are visibly removed if inpainting is applied. Additionally, the overall disparity values are closer to the ground truth with inpainting.

These disparity results were confirmed by the quantitative evaluation in Table \ref{tab:disparity}. We measure the root mean square error (RMS), the endpoint error (EPE), and the percentage of the disparity image with more than three pixel disparity error (bad3\% error). The errors are relative to ground truth disparities obtained from CT scanning and Creaform RGB scanning, as indicated in the table. The evaluations were also done for occluded and non-occluded pixels separately. The difference between the RMS and EPE of the original versus the inpainted-based disparity estimations were all positive meaning that there was an improvement in the disparity estimation due to the inpainting. The Bad3\% metric is better in Experiment 2, but not in Experiment 1 which indicates that our overall improvements in this particular case are mostly due to reducing higher errors and extreme outliers. 

\begin{table}[t!]
    \centering
    \begin{tabular}{|c|c|c|c|c|c|}
    \hline
    Exp.   & Modality & Occ. &	Bad3     &	RMS   &	EPE\\
    \hline
    \hline
    1 &	CT      &	0 &-0.00107  &	0.085 &	0.0552\\
    \hline
    1 &	CT      &	1 &-0.00107  &	0.085 &	0.0554\\
    \hline
    2 &	CT      &	0 &0.00310   &	0.102 &	0.0842\\
    \hline
    2 &	CT      &	1 &0.00235   &	0.100 &	0.0820\\
    \hline
    2 &	RGB     &	0 &0.00288   &	0.103 &	0.0838\\
    \hline
    2 &	RGB     &	1 &0.00289   &	0.102 &	0.0820\\
    \hline
    \end{tabular}
    \caption{The difference between quantitative evaluations of disparity generation from original frames, $Disp_{orig}$, and that from inpainted frames using $Model_{T,C}$, $Disp_{inp}$. The tabulated values are those of $Disp_{orig}$ - $Disp_{inp}$. Two experiments (Exp.) are used with different ground truth modes (Modality) and with and without occluded pixels (Occ.)}
    \label{tab:disparity}
\end{table}

\subsection{Sparse Feature Matching}

\begin{figure}[t!]
    \centering
    \subfigure[Original Frame Matches]{\includegraphics[width=0.48\columnwidth]{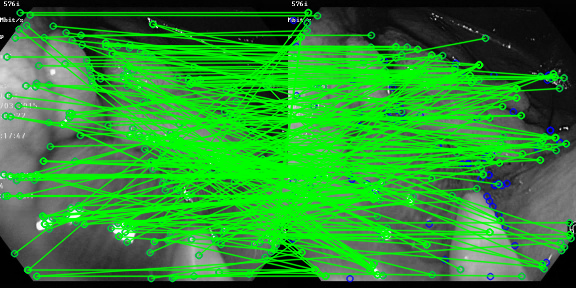}}
    \subfigure[Inpainted Frame Matches]{\includegraphics[width=0.48\columnwidth]{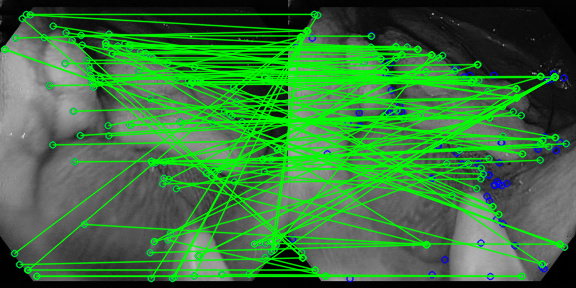}}
     \subfigure[Original Frame Filtered Matches]{\includegraphics[width=0.48\columnwidth]{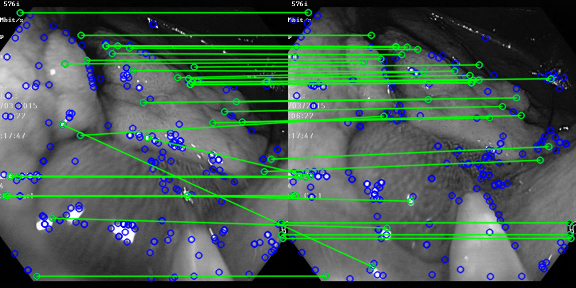}}
     \subfigure[Inpainted Frame Filtered Matches]{\includegraphics[width=0.48\columnwidth]{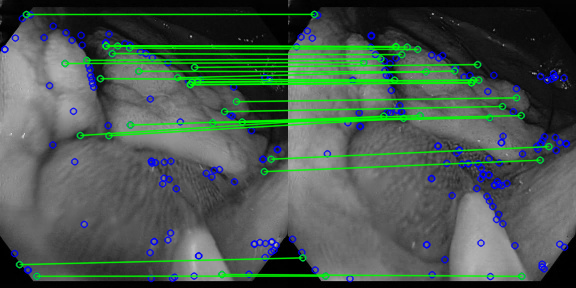}}

    \caption{Matches between features are shown in Green and the removed low quality features are shown in Blue. It can be seen that the original frame pair matches are much more than those generated from the inpainted frame pair. This indicates that without inpainting specular highlights, more lower quality feature matches are generated. }
    \label{fig:FM}
\end{figure}

To assess the effect of our method on sparse feature matching, we detect and match ORB features \citep{rublee2011orb} on the SCARED dataset. Qualitative results can be visualised in Fig. \ref{fig:FM}. Inpaining reduces the number of detected features in image pairs, but these are mostly features located on specular highlights that have high gradients but are generally outliers. After outlier filtering using a ratio test, the number of detected matches is similar.

\begin{table*}[t!]
\small
    \centering
    \begin{tabular}{c|c|c|c|c|c|c|c|c|c}
 Statistics       & Mean(Orig)& Mean(Inp) & Mean($\Delta$)   & Min($\Delta$)& Max($\Delta$)& $25^{th}(\Delta)$& Median($\Delta$) & $75^{th}$($\Delta$) & IQR($\Delta$) \\
 \hline
 RTE ($^{\circ}$) & 65.59     & 57.29     & 8.31  (\% 11.72) & -139.16      & 165.06       & -11.39           & 4.69             & 28.68               & 40.07\\
 RRE ($^{\circ}$) & 6.64      & 4.64      & 2.0   (\% 30.12) & -168.69      & 175.66       & -0.45            & 0.06             & 0.71                & 1.16\\
 Inliers (pixels) & 90.12     & 67.45     & 22.68 (\% 25.17) & -221.0       & 240.0        & -5.5             & 22.0             & 51.5                & 57.0
    \end{tabular}
    \caption{Statistics summary (Mininum, Maximum, Mean, $25^{th}$ percentile, Median, $75^{th}$ percentile, and Interquartile Range (IQR)) for the difference ($\Delta$) between the feature-based pose estimation results of original versus inpainted sequences in terms of RTE, RRE and Inliers. This shows the positive effect of inpainting on feature matching.}
    \label{tab:statbox}
\end{table*}

\begin{figure*}[t!]
    \centering
    \subfigure[]{\includegraphics[width=0.3\textwidth]{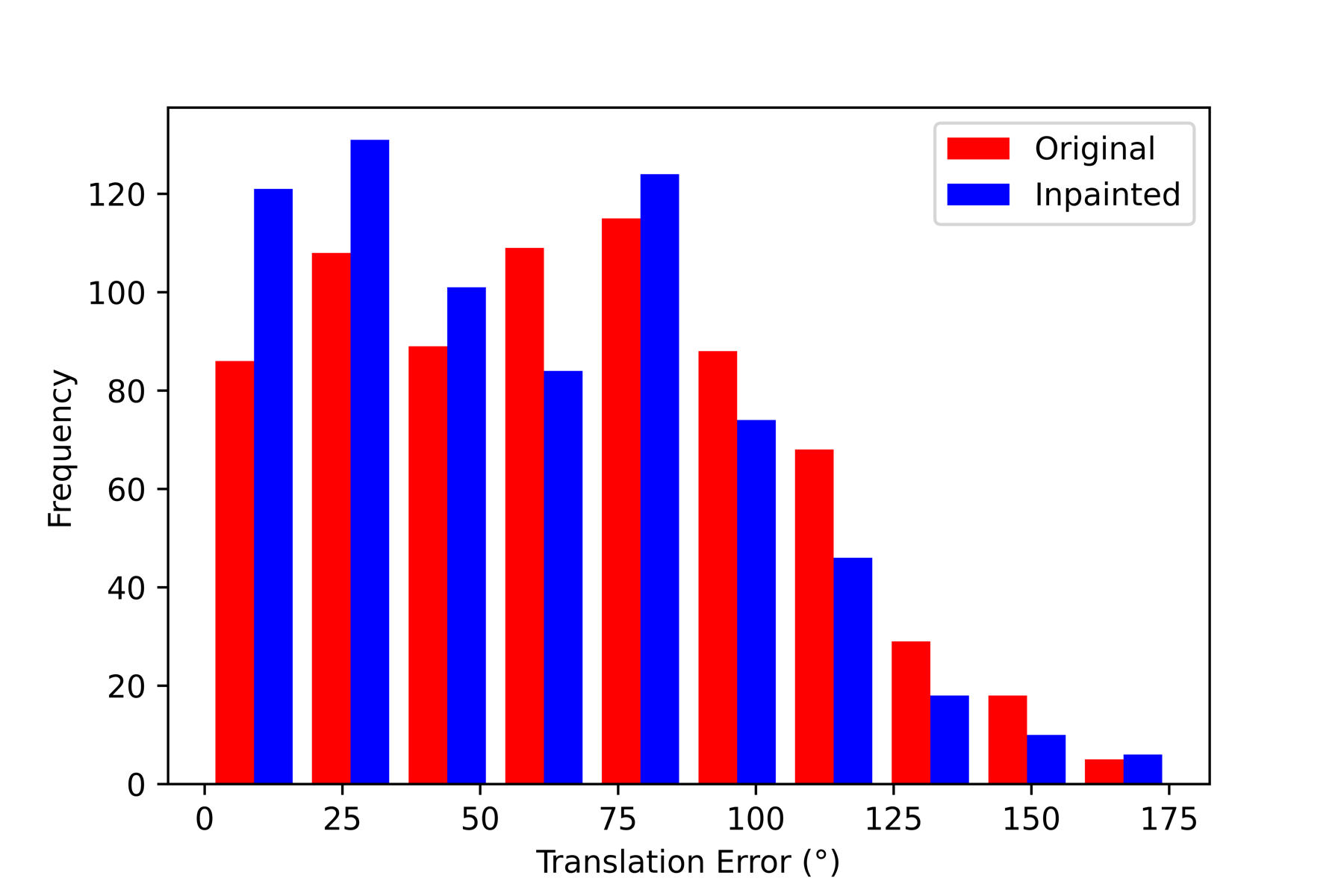}}
    \subfigure[]{\includegraphics[width=0.3\textwidth]{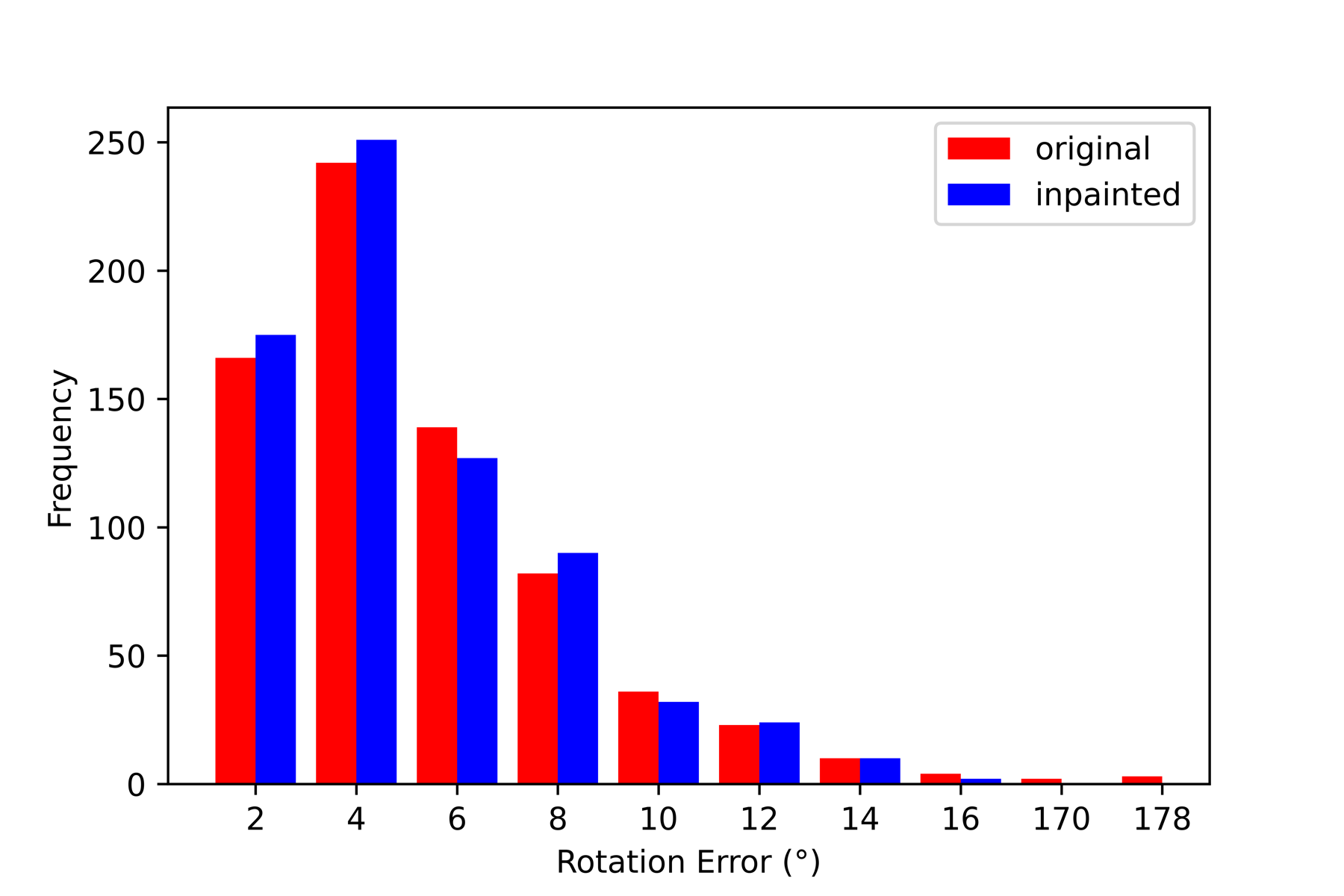}}
    \subfigure[]{\includegraphics[width=0.3\textwidth]{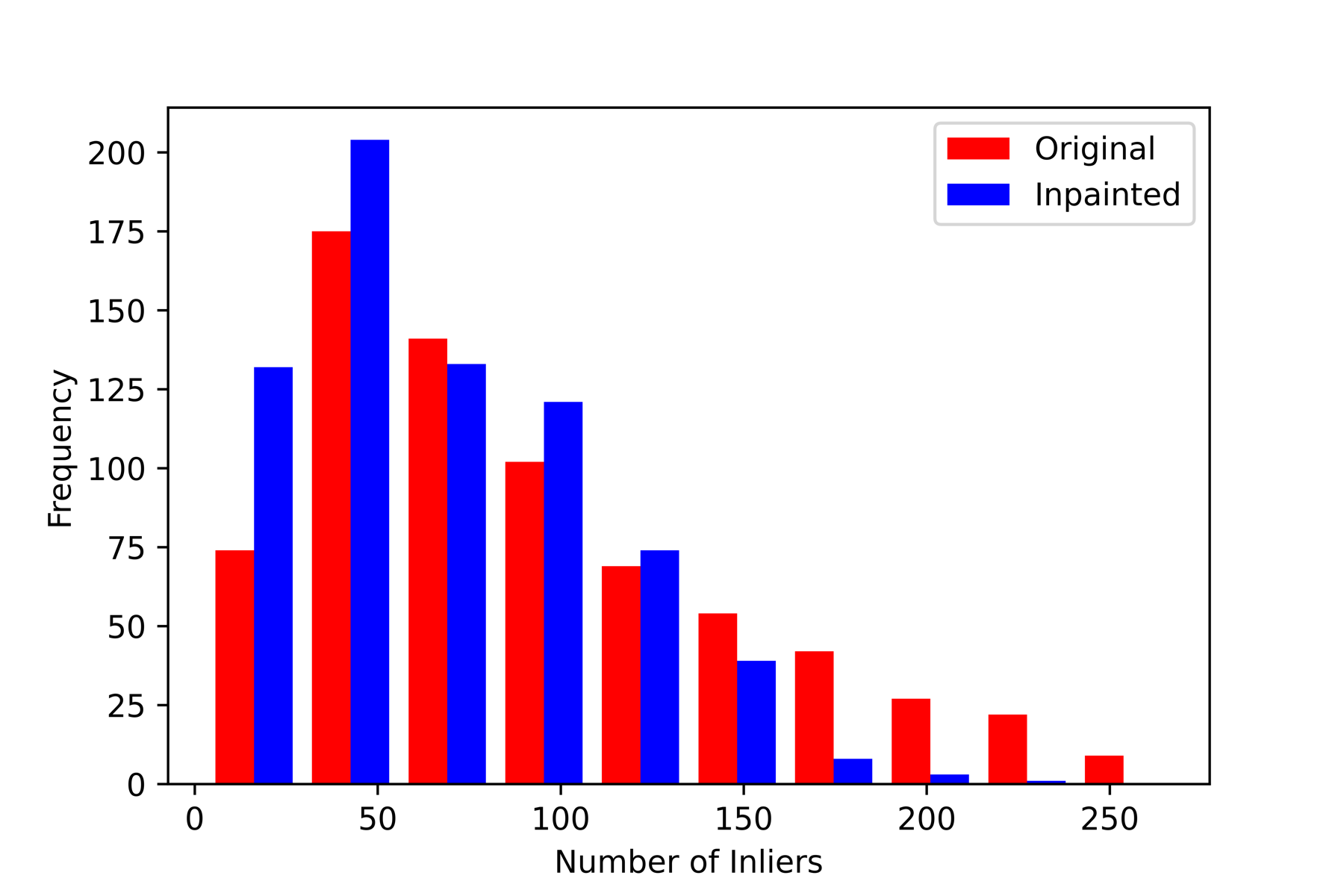}}
    \caption{Feature Matching Analysis: Histograms of RTE, RRE, and RANSAC Inliers for both original and inpainted sequences. It can be seen that the inpainted frames have more low errors than the original frames. This shows the improvement of feature matching and thus pose estimation due to inpainting.}
    \label{fig:RTE}
\end{figure*}

\begin{figure*}[t!]
    \centering
    \subfigure[Original sequence]{\includegraphics[width=0.4\textwidth]{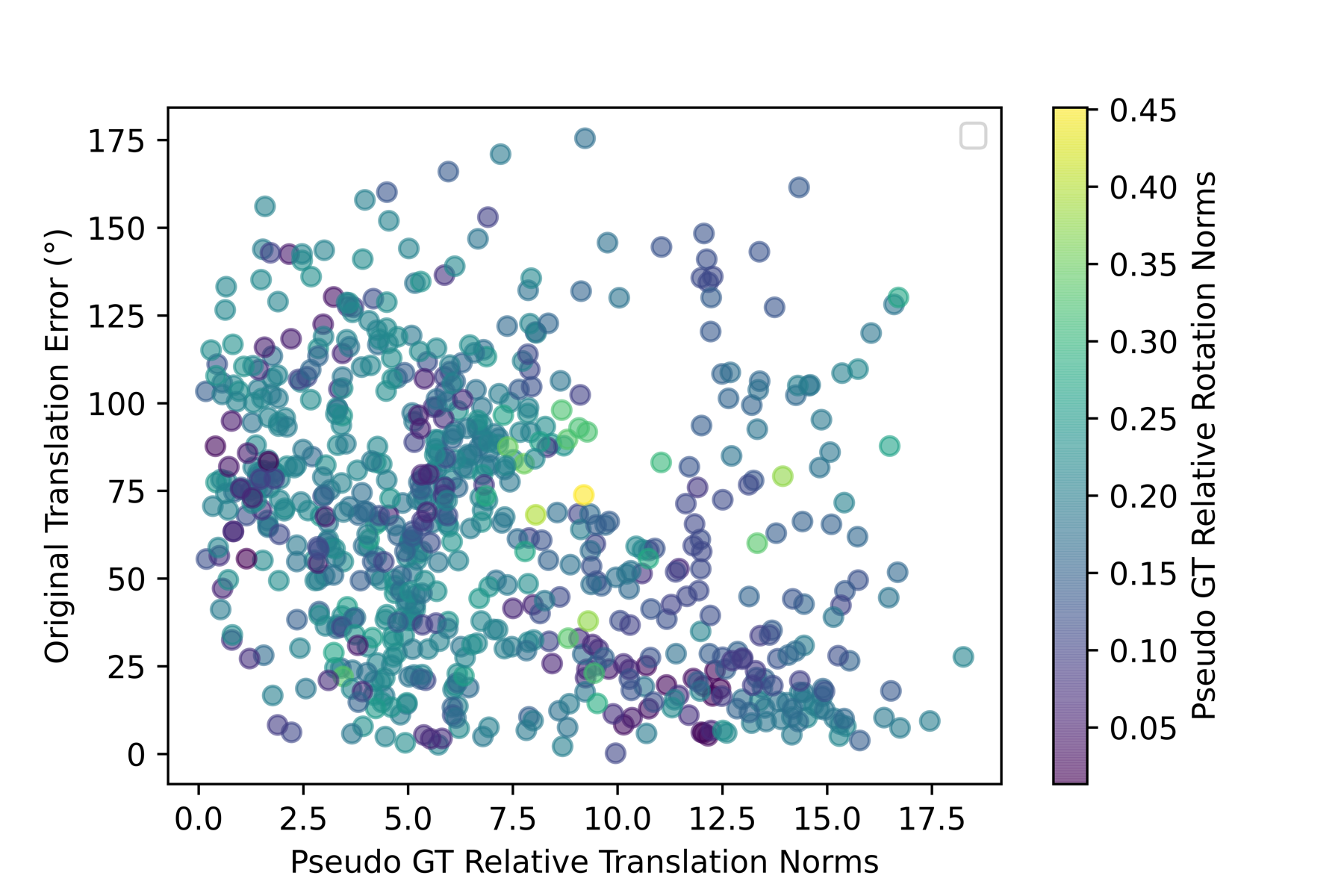}}
    \subfigure[Inpainted sequence]{\includegraphics[width=0.4\textwidth]{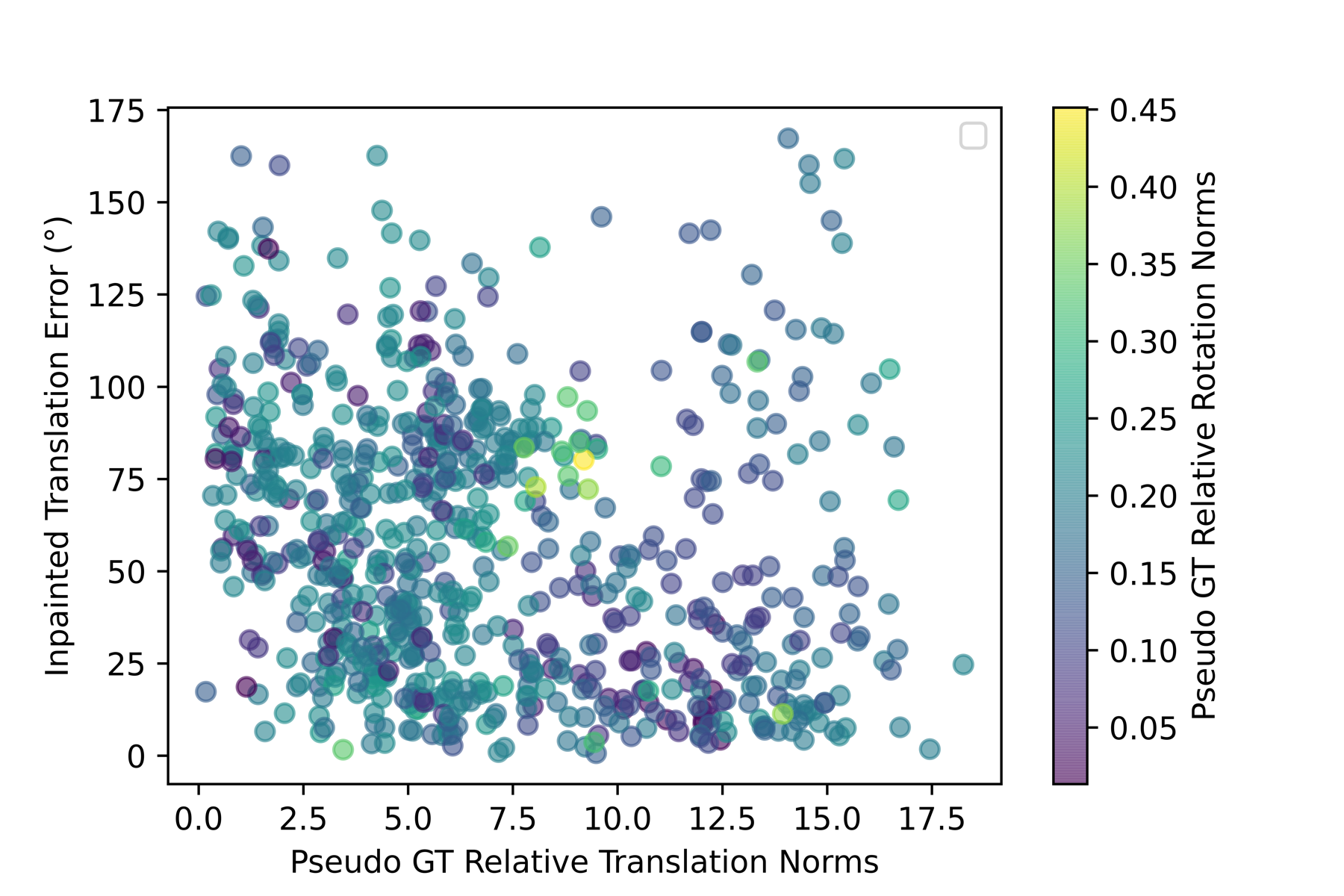}}
    \caption{Effect of the ground truth translation and rotation norms on RTE. Low translation norms result in high errors or noise. }
    \label{fig:RTEeffect}
\end{figure*}

To assess the accuracy of these matches, we use them for estimating relative poses between pairs of camera views. Ground truth of camera relative poses is available via the captured robot kinematics as part of the SCARED dataset. Relative pose is estimated with the 5-point algorithm solver within the RANSAC robust estimator \citep{nister2004efficient}. The relative translation and rotation errors (RTE and RRE in degrees) as well as the number of inliers detected by RANSAC were calculated with respect to the ground truth. 

A statistical summary of the RTE, RRE, and number of inliers is shown in Table \ref{tab:statbox}. For RTE, the interquartile range ($[-11.39,28.68]^{\circ}$) of the difference ($\Delta$) between original and inpainted sequences is larger on the positive region and the median ($4.69^{\circ}$) and mean (\% 11.72) fall completely in the positive region. The positive numbers mean that the RTEs for the original sequence are higher than those for the inpainted sequence, which shows that inpainting can help in increasing accuracy of relative translation estimation, as well as providing a higher number of sparse features that are inliers after geometric verification, which can be useful for reconstruction and visual SLAM methods.

One can also notice that the mean RTE values are very high in general. After investigating and plotting the effect of the relative translation and rotation norms on the RTE as shown in Fig. \ref{fig:RTEeffect} (a) (b), a clear effect can be seen between the relative translation norms and the RTE. This is a well known problem in relative pose estimation with monocular cameras, where errors of small translations are difficult to assess and may be under the precision of robot kinematics used to generate ground truth. 

A similar analysis can be done for the relative rotation error (RRE). However, the RREs are reasonably small with a few outliers. We can see the skewness of the interquartile range ($[-0.45, 0.71]^{\circ}$), median ($0.06^{\circ}$), and mean (\% 30.12) to the positive region in Table \ref{tab:statbox}, which indicates an improvement in relative rotation estimation from feature matching after inpainting.

This improvement is also shown in Fig. \ref{fig:RTE} (b), where the inpainted sequence has lower errors (0-25) than the original. Note that for clarity in Fig. \ref{fig:RTE} (b), all bars with height less or equal to 1 are removed.

Finally, the numbers in Table \ref{tab:statbox} show that the original sequence has a higher number of inliers than the inpainted one with an interquartile range ($[-5.5, 51.5] pixels$), which is much larger on the positive region, and a median (22.0 pixels) and mean (\% 25.17) falling completely in the positive region. This can indicate that fewer pixels are wrongly identified as inliers. 

This improvement is also shown in Fig. \ref{fig:RTE} (c), where the inpainted sequence has more lower number of inliers (0-125) and less of the higher number of inliers (150-250) than the original.

To conclude, feature matching was improved through our inpainting method. This was evaluated visually as well as quantitatively through pose estimation and the essessment of RTE, RRE, and RANSAC inliers.

\subsection{Optical Flow} 

\begin{figure*}[t!]
    \centering
    \subfigure[]{\includegraphics[height=0.2\textheight]{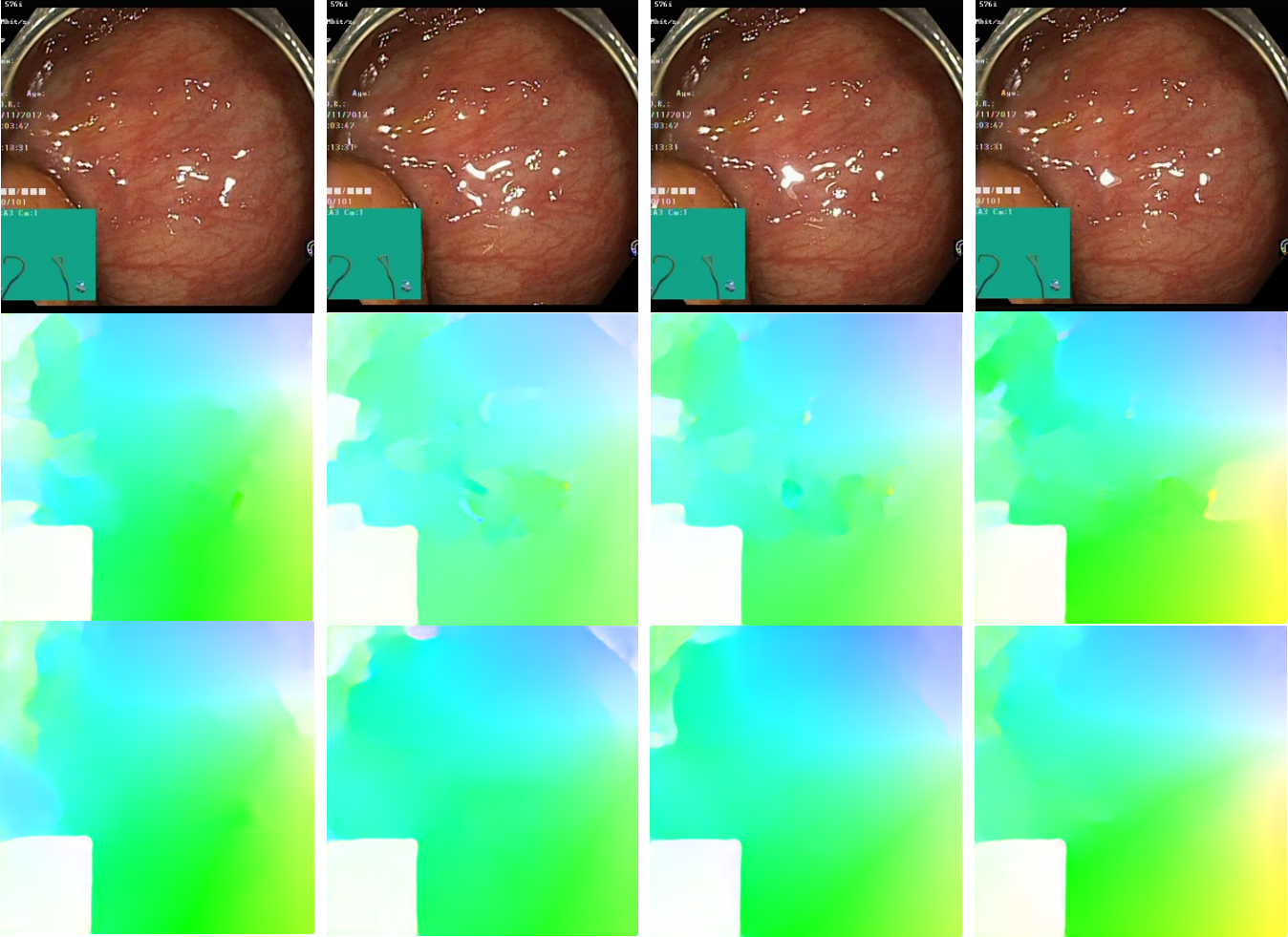}}
    \subfigure[]{\includegraphics[height=0.2\textheight]{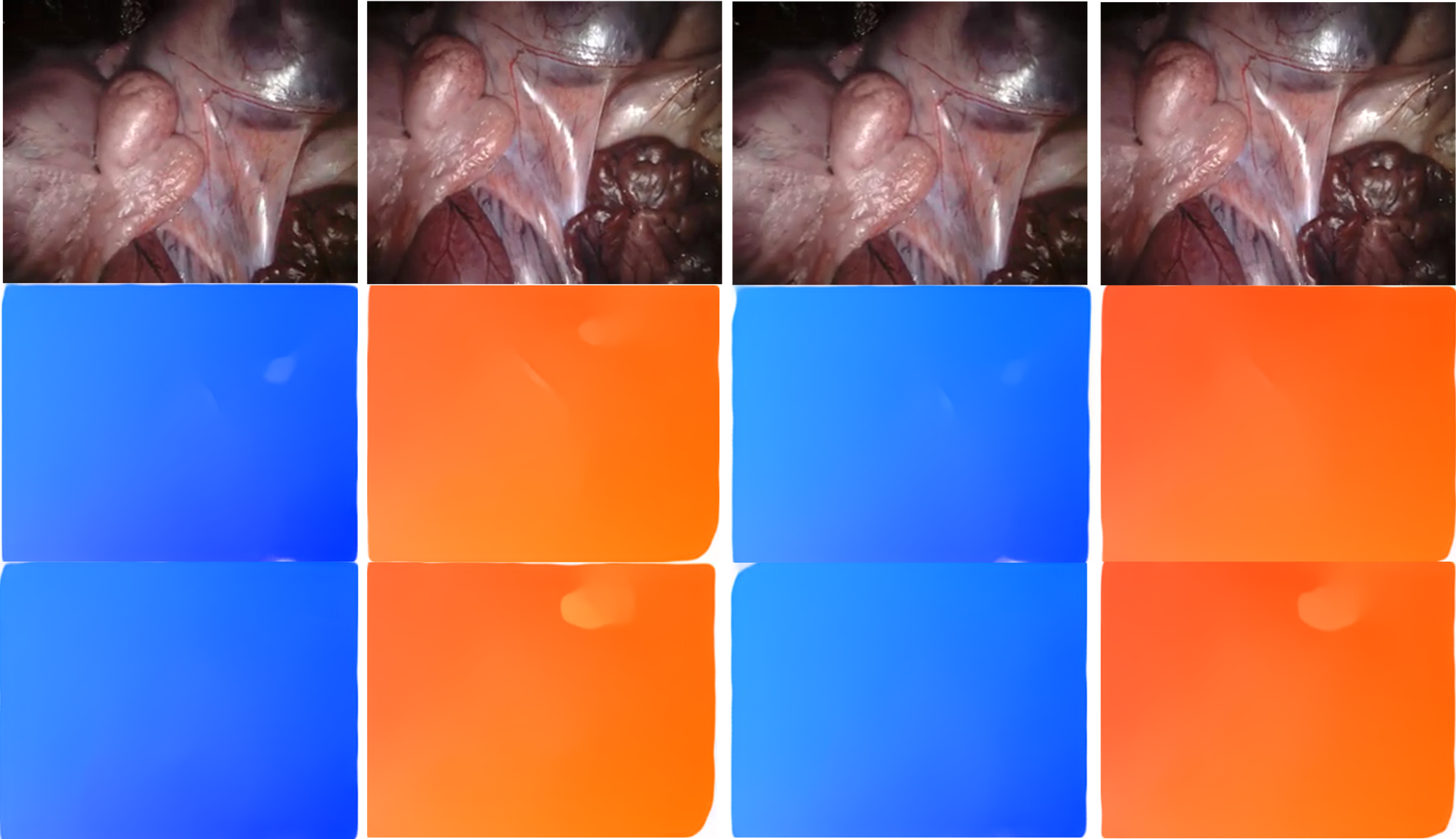}}
    \subfigure[]{\includegraphics[width=0.3\columnwidth]{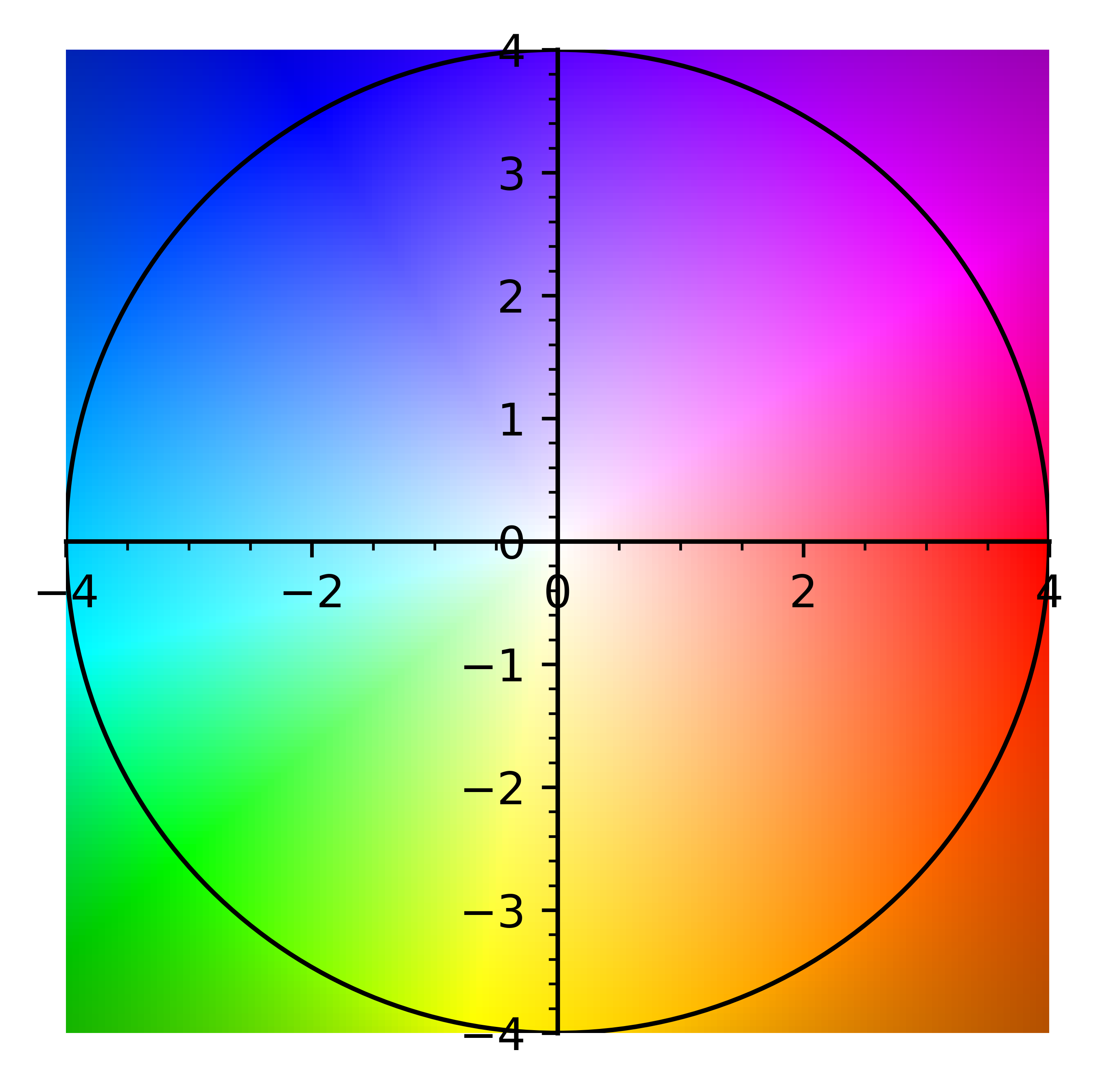}}
    \caption{(a) From four consecutive frames of the Hyper-Kvasir dataset (First Row), optical flow is estimated directly (Second Row) and after inpainting the frames using $Model_{T,C}$ (Third Row). Inpainting improved the flow estimation results by making it smoother and more homogeneous. (b) From four consecutive frames of the SCARED dataset (First Row), optical flow is estimated directly (Second Row) and after inpainting the frames using $Model_{T,C}$ (Third Row). It is is not clear if inpainting improved the flow estimation results. The SCARED dataset used differs from the training dataset (Hyper-Kvasir) in its surgical procedure performed. (c) Optical flow color wheel.}
    \label{fig:OF2}
\end{figure*}

The effect of our method on optical flow, using FLOWNET2.0, can be qualitatively visualized in Fig. \ref{fig:OF2} for Hyper-Kvasir and SCARED datasets. It can be seen that the optical flow generated from the inpainted frames is smoother, and has fewer spurious artifacts. More results are provided in the supplementary material. 

Similar to the previous section, we evaluate the optical flow results quantitatively via relative pose estimation between pairs of frames, using the same algorithmic and validation pipeline.

The RTE, RRE, and number of inliers were statistically summarized in Table \ref{tab:statboxOF}. The interquartile range ($[-8.11,10.52]^{\circ}$) of the difference ($\Delta$) in RTE between original and inpainted is skewed to the positive region. In addition, the mean is + \% 1.73, meaning that inpainting increased translation estimation accuracy by \% 1.73 on average, which is a very small number and is not enough to show improvement. 

\begin{table*}[t!]
\small
    \centering
    \begin{tabular}{c|c|c|c|c|c|c|c|c|c}
 Statistics       & Mean(Orig)& Mean(Inp) & Mean($\Delta$)     & Min($\Delta$)& Max($\Delta$)& $25^{th}(\Delta)$& Median($\Delta$) & $75^{th}$($\Delta$) & IQR($\Delta$) \\
 \hline
 RTE ($^{\circ}$) & 38.20     & 37.54     & 0.67 (\% 1.73)     & -141.72      & 122.37       & -8.11            & -0.19            & 10.52               & 18.63\\
 RRE ($^{\circ}$) & 3.14      & 3.19      & -0.05 (-\% 1.59)   & -8.60        & 8.83         & -0.68            & -0.06            & 0.56                & 1.24\\
 Inliers (pixels) & 50457     & 52890     & -2432 (-\% 4.82)   & -61971       & 61287        & -4922            & -126             & 2789            & 7711
    \end{tabular}
    \caption{Statistics summary (Mininum, Maximum, Mean, $25^{th}$ percentile, Median, $75^{th}$ percentile, and Interquartile Range (IQR)) for the box plots of the difference between the optica-flow-based pose estimation results of original versus inpainted sequences in terms of RTE, RRE and Inliers. This shows the low effect of inpainting on optical flow.}
    \label{tab:statboxOF}
\end{table*}

\begin{figure*}[t!]
    \centering
    \subfigure[]{\includegraphics[width=0.3\textwidth]{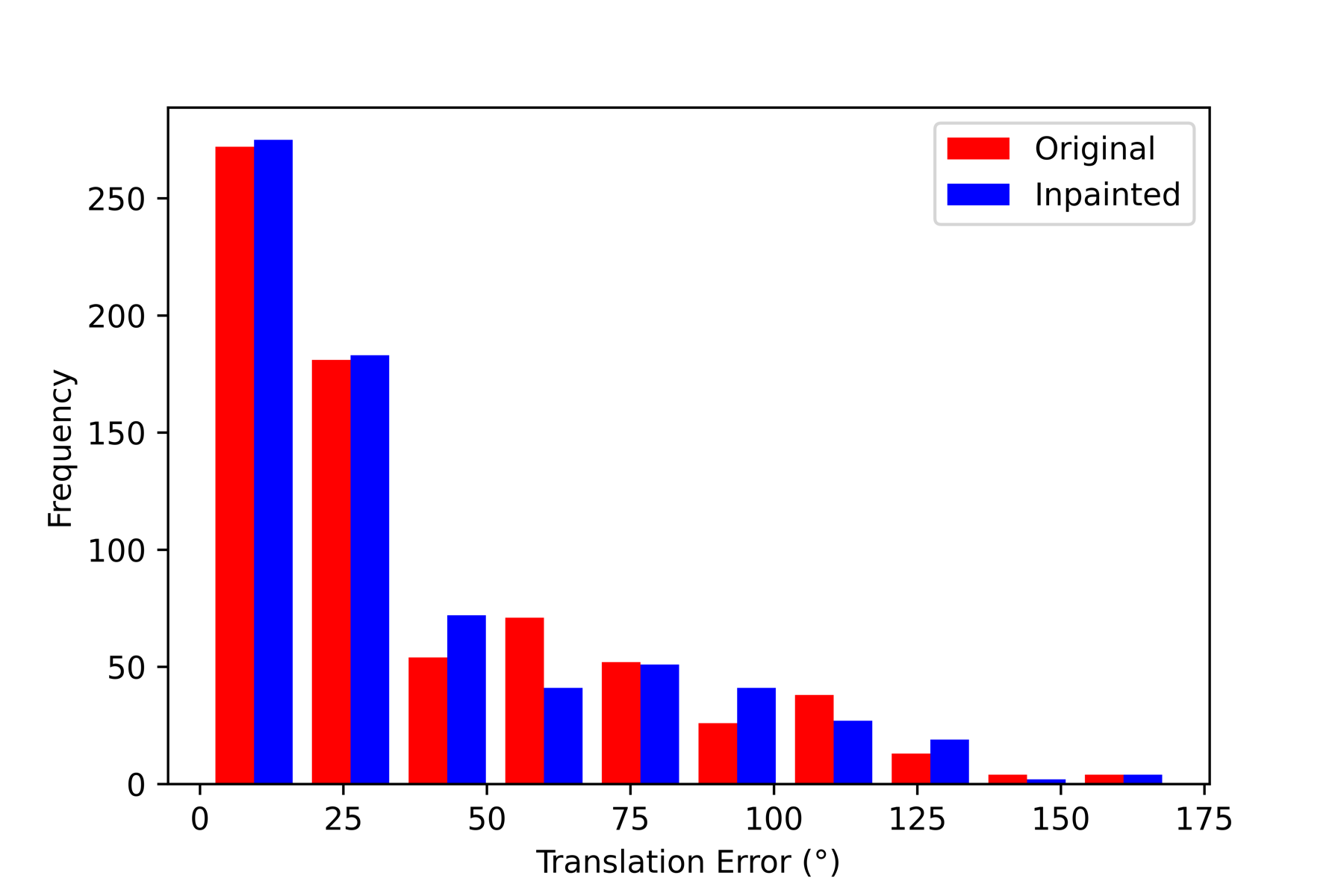}}
    \subfigure[]{\includegraphics[width=0.3\textwidth]{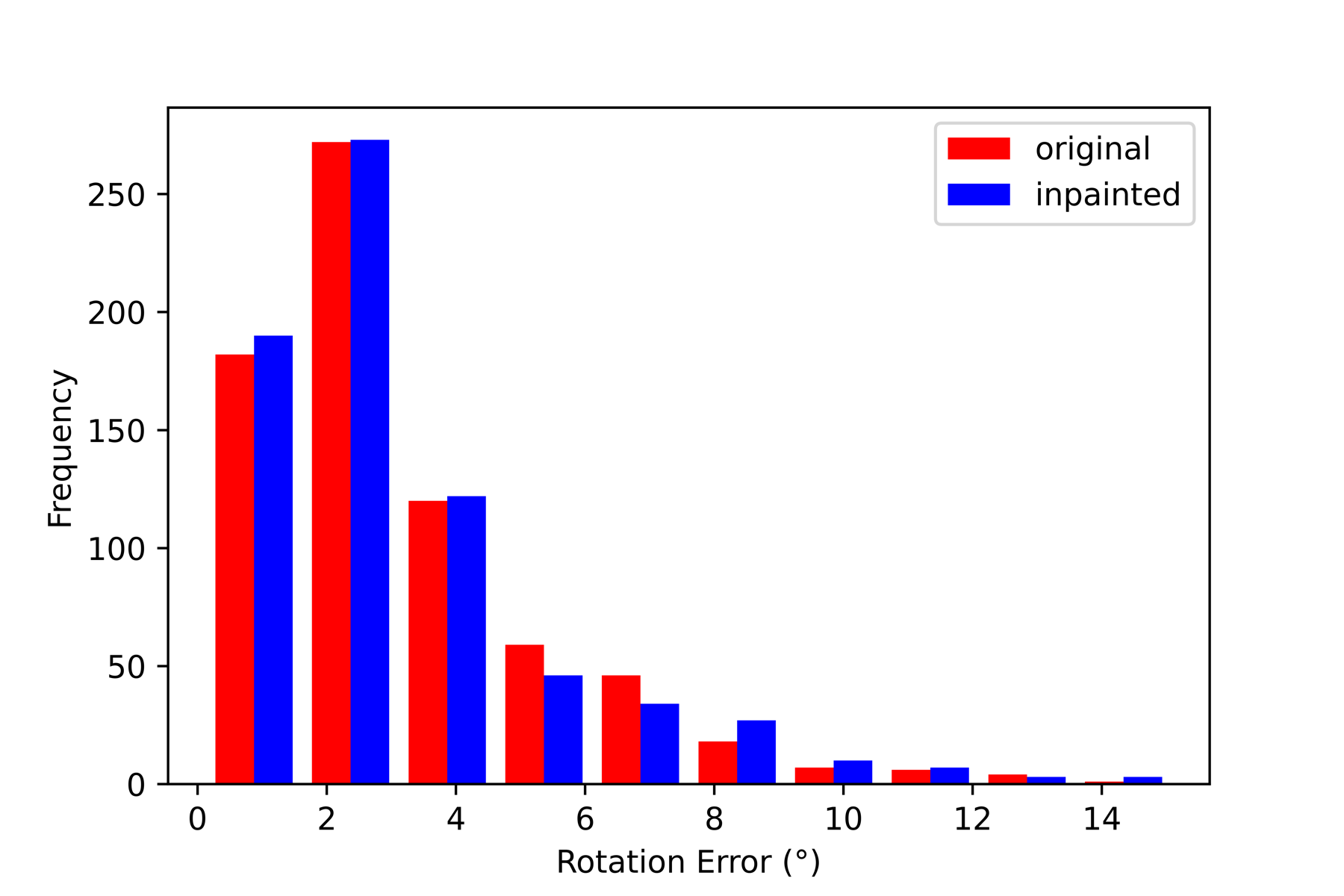}}
    \subfigure[]{\includegraphics[width=0.3\textwidth]{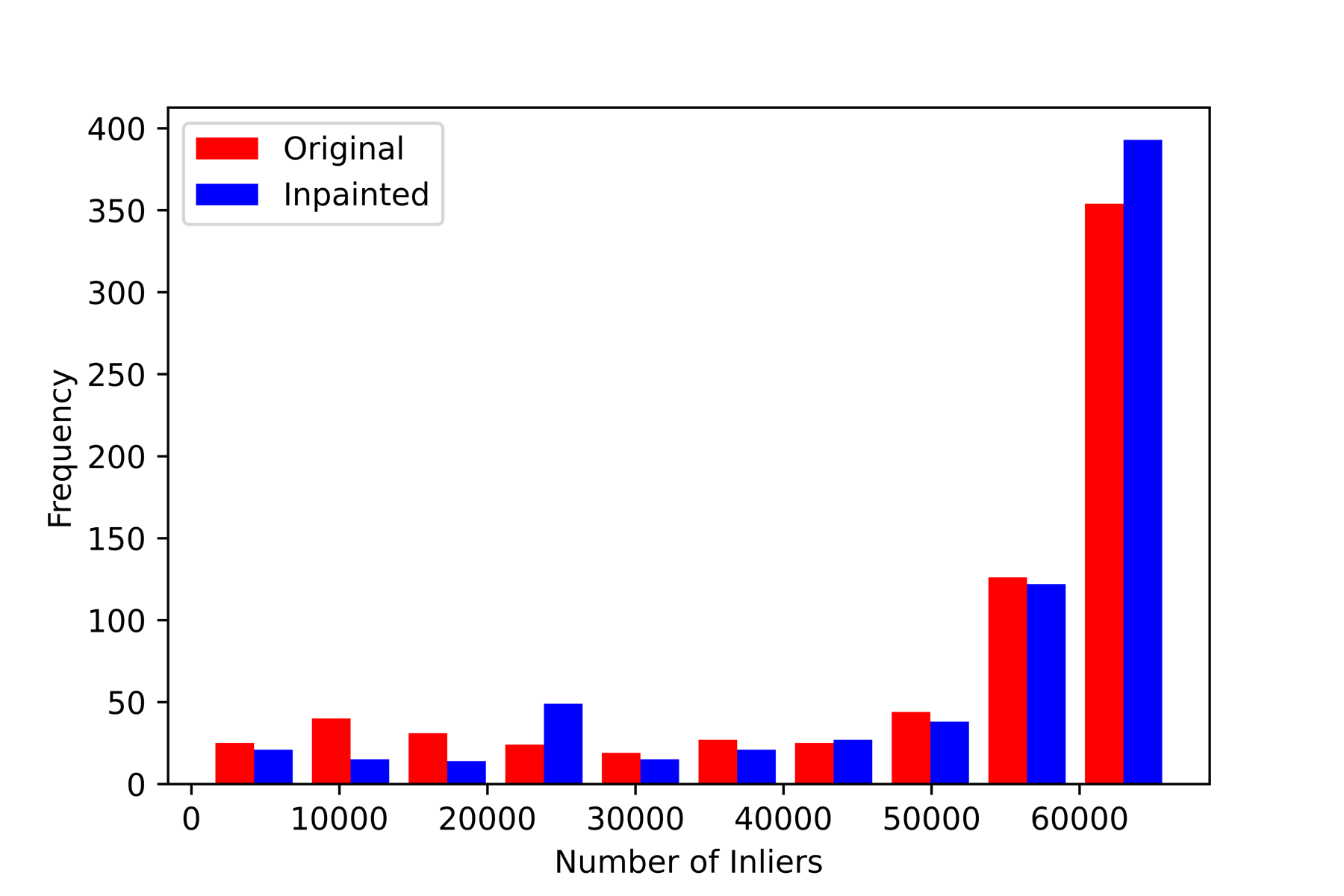}}
    \caption{Optical Flow Analysis: Histograms of RTE, RRE, and RANSAC Inliers for both original and inpainted sequences. No significant changes can be seen in these plots. This shows that optical flow is not improved for the SCARED dataset when inpainted.}
    \label{fig:RTEOF}
\end{figure*}

To further visualize these numbers, looking at Fig. \ref{fig:RTEOF} (a), we can see that the inpainted sequence has slightly lower errors (0-50) then the original and no apparent increase or decrease in the higher errors (50-175).

\begin{figure*}[t!]
    \centering
    \subfigure[Original sequence]{\includegraphics[width=0.4\textwidth]{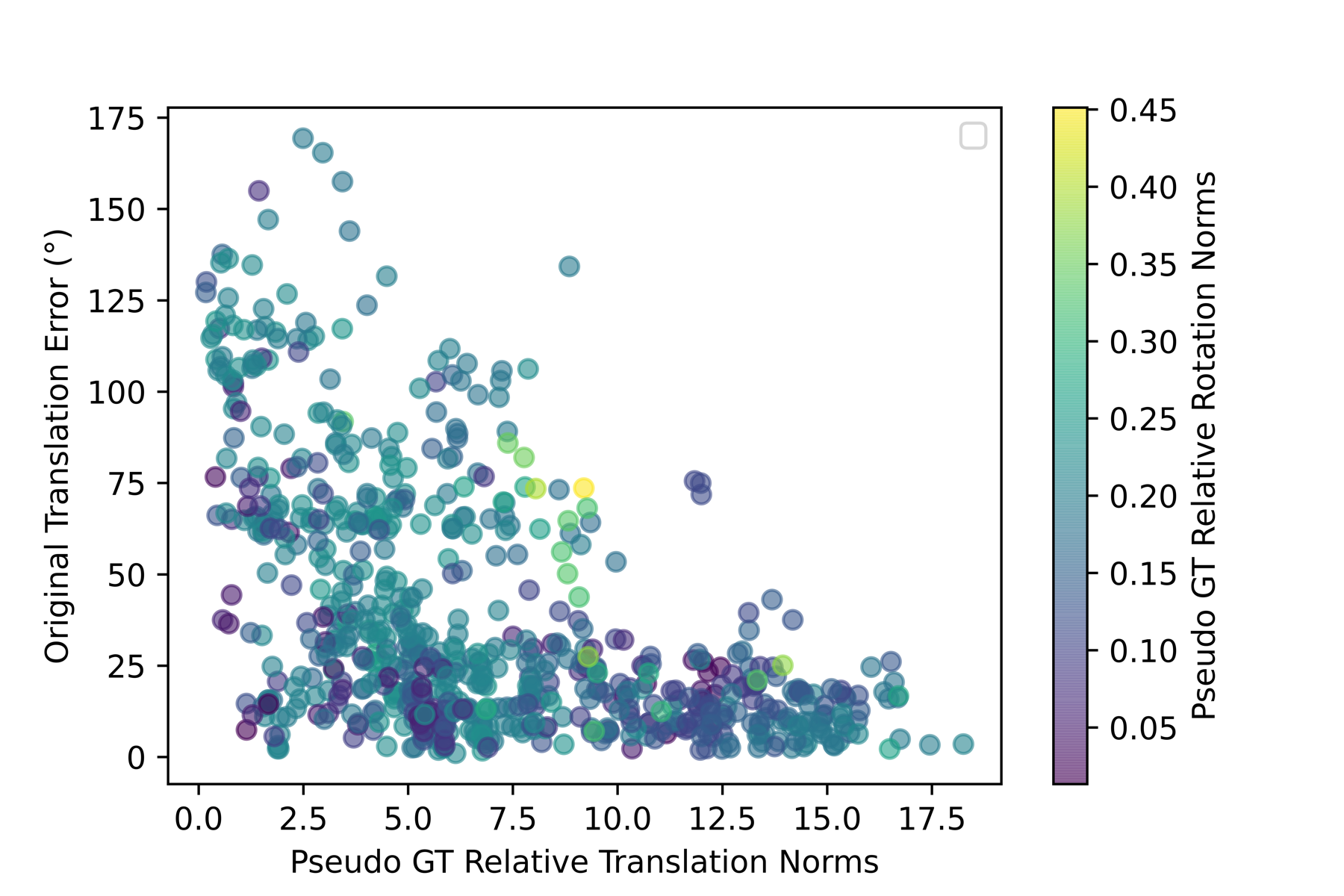}}
    \subfigure[Inpainted sequence]{\includegraphics[width=0.4\textwidth]{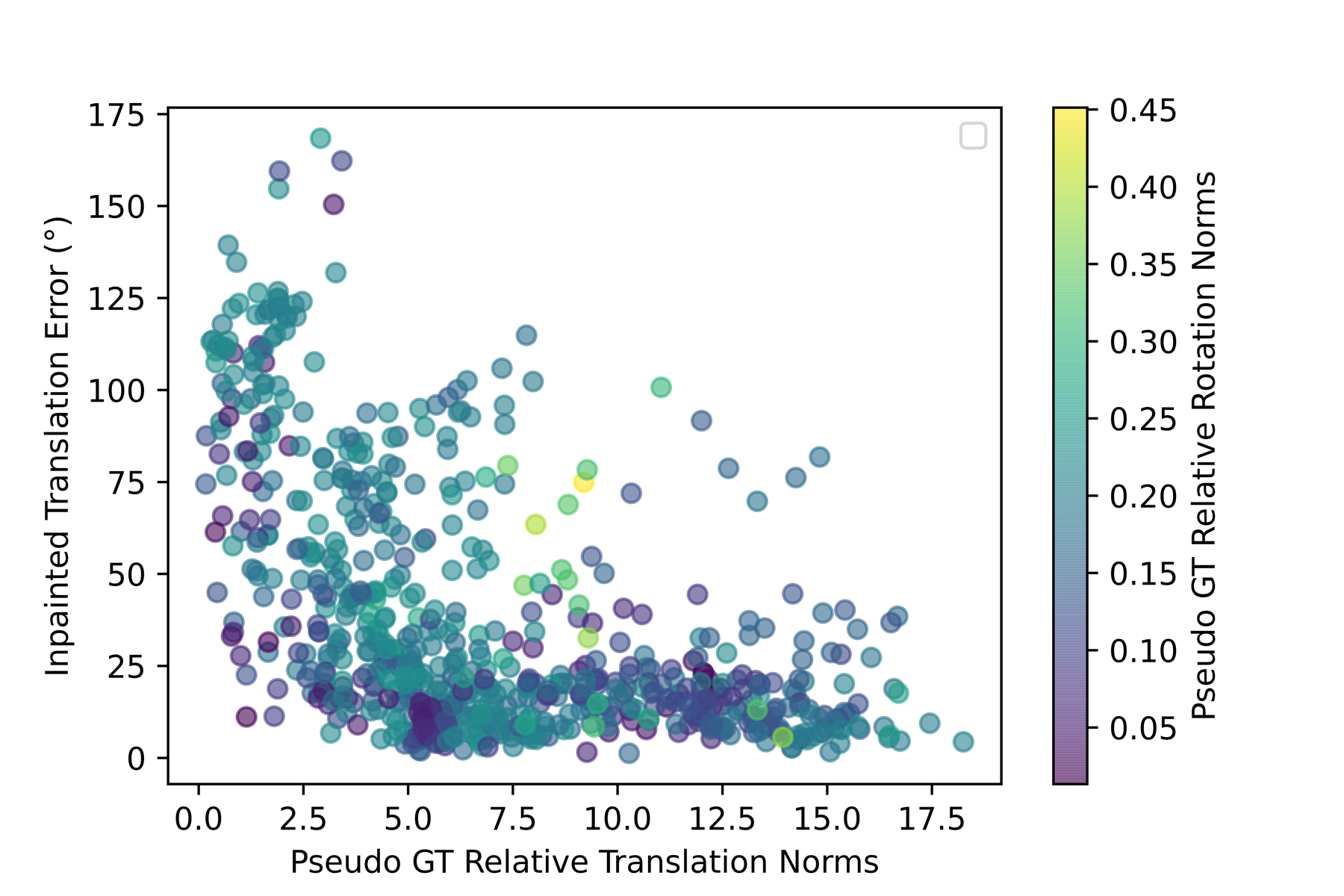}}
    \caption{Effect of the ground truth translation and rotation norms on RTE. Low translation norms result in high errors or noise.}
    \label{fig:RTEeffectOF}
\end{figure*}

From Table \ref{tab:statboxOF}, the RTE and RRE improvements due to inpainting are not as high as those with feature matching. However, mean RTE values are still quite high. Similar to feature matching, the relationship between translation errors and translation norms shows a high correlation in Fig. \ref{fig:RTEeffectOF}. 

Similarly, for relative rotation errors we look at the $\Delta$ RRE statistics in Table \ref{tab:statboxOF}, where the interquartile range ($[-0.68,0.56]^{\circ}$), mean (- \%1.59), and median ($-0.06^{\circ}$) between original and inpainted are all skewed to the negative region. This means our method does not improve the overall motion estimation accuracy with optical flow. This is notoriously different from sparse feature matching. It can be explained by optical flow providing dense correspondences, providing a surplus of information when compared to sparse features, that makes RANSAC more effective at removing outliers caused by spurious artifacts. We also note that our method has not been fine-tuned on laparoscopic data (SCARED dataset), and therefore further improvements could be obtained with additional surgery-specific training data. This is consistent with the more visible qualitative improvements on the Hyper-Kvasir data, which is in the same domain as our training data.

\section{Limitations}

The main limitation of the proposed system is its reliance on the detection method. This can sometimes miss some specularities and be thrown off by text and diagram information overlayed by endoscopic camera interfaces. This can be seen for example in Fig. \ref{fig:masks}, where the text and green square are detected as specularities and are later on inpainted. This can affect the learning process of the temporal GAN and limit its ability to find similarity between occluded regions of different frames. In the future, the system can be expanded to include a detection system of static screen objects and remove them from consideration. 

\section{Conclusion}
In this paper, a system was proposed to inpaint specular highlights in MISD videos. An endoscopic pseudo ground truth dataset was generated after which a model was trained using temporal GANs and transfer learning to produce inpainted frames based on spatio-temporal information. The qualitative and quantitative results of the system showed improvement on traditional methods. An ablation study was also carried out to show the importance of the temporal component and transfer learning in generating enhanced frames. In addition, the effect of this system on various applications including feature matching, disparity estimation, and optical flow prediction was shown to be significant. In the future, a more precise detection pipeline can be introduced as well as an extension of this research to other artefacts and to the 3D reconstruction world.

\section*{Acknowledgments}
This work was supported by the Wellcome/EPSRC Centre for Interventional and Surgical Sciences (WEISS) at UCL(203145Z/16/Z) and H2020 FET(GA863146).
\bibliographystyle{model2-names.bst}\biboptions{authoryear}
\bibliography{refs}

\end{document}